RESEARCH ARTICLE

# TUM autonomous motorsport: An autonomous racing software for the Indy Autonomous Challenge

Johannes Betz[1] | Tobias Betz[1] | Felix Fent[1] | Maximilian Geisslinger[1] | Alexander Heilmeier[1] | Leonhard Hermansdorfer[1] | Thomas Herrmann[1] | Sebastian Huch[1] | Phillip Karle[1] | Markus Lienkamp[1] | Boris Lohmann[2] | Felix Nobis[1] | Levent Ögretmen[2] | Matthias Rowold[2] | Florian Sauerbeck[1] | Tim Stahl[1] | Rainer Trauth[1] | Frederik Werner[1] | Alexander Wischnewski[2]

[1]Technical University of Munich, School of Engineering & Design, Institute of Automotive Technology (FTM), Garching, Germany

[2]Technical University of Munich, School of Engineering & Design, Chair of Automatic Control (RT), Garching, Germany

**Correspondence**
Phillip Karle, Technical University of Munich, School of Engineering & Design, Institute of Automotive Technology (FTM), Boltzmannstr. 15, Garching 85748, Germany.
Email: phillip.karle@tum.de

## Abstract

For decades, motorsport has been an incubator for innovations in the automotive sector and brought forth systems, like, disk brakes or rearview mirrors. Autonomous racing series such as Roborace, F1Tenth, or the Indy Autonomous Challenge (IAC) are envisioned as playing a similar role within the autonomous vehicle sector, serving as a proving ground for new technology at the limits of the autonomous systems capabilities. This paper outlines the software stack and approach of the *TUM Autonomous Motorsport* team for their participation in the IAC, which holds two competitions: A single-vehicle competition on the Indianapolis Motor Speedway and a passing competition at the Las Vegas Motor Speedway. Nine university teams used an identical vehicle platform: A modified Indy Lights chassis equipped with sensors, a computing platform, and actuators. All the teams developed different algorithms for object detection, localization, planning, prediction, and control of the race cars. The team from Technical University of Munich (TUM) placed first in Indianapolis and secured second place in Las Vegas. During the final of the passing competition, the TUM team reached speeds and accelerations close to the limit of the vehicle, peaking at around 270 $km\,h^{-1}$ and 28 $ms^{-2}$. This paper will present details of the vehicle hardware platform, the developed algorithms, and the workflow to test and enhance the software applied during the 2-year project. We derive deep insights into the autonomous vehicle's behavior at high speed and high acceleration by providing a detailed competition analysis. On the basis of this, we deduce a list of lessons learned and provide insights on promising areas of future work based on the real-world evaluation of the displayed concepts.

All authors contributed equally.

# 1 | INTRODUCTION

## 1.1 | Motivation

Racing has been a platform for innovation since its very beginning. Safety mechanisms, powertrain, and suspension technology as well as tires have been improved during the past decades in several competition formats. Recently, autonomous racing became a proving ground for autonomous vehicle technology at the limits of its current capabilities. The most prominent examples include the F1Tenth racing series, Formula Student Driverless (FSD), Roborace, and the Indy Autonomous Challenge (IAC). While each of those series has a slightly different scope and focus, all of them target the improvement of the used sensors, actuators, and compute platforms as well as the development of the required algorithms, middleware, and operating systems. The race track provides a safe proving ground for high-speed testing and challenges autonomous vehicles frequently with complex scenarios.

A research team from the Technical University of Munich (TUM) decided to participate in the IAC (Figure 1a) and the follow-up event, the Autonomous Challenge at CES in Las Vegas (AC@CES, Figure 1b), in October 2021 and January 2022. Nine teams from international universities took part in the real-world events and competed in two different formats: First, the target in Indianapolis was a combination of setting the fastest lap on the Indianapolis Motor Speedway (IMS) and demonstrating dynamic obstacle evasion capability. Second, the event in Las Vegas was based on a head-to-head passing competition with alternating overtaking attempts of two participants with increasing speeds for each round. The TUM team finished first at the inaugural event at the IMS and second at the AC@CES.

The competition focused solely on the development of the required autonomous racing software stack. Therefore, all the vehicles were based on the same chassis as well as the same sensors, actuators, and compute platforms. It started initially with 31 teams and a series of hackathons built around simulated racing challenges with increasing complexity, leading up to multiple eight-vehicle simulation races in May 2021. Finally, nine teams were asked to deploy their software on the vehicles starting in July 2021 and practiced on a small oval race track, Lucas Oil Raceway, before moving to the larger IMS and the Las Vegas Motor Speedway.

This paper introduces the approach of the *TUM Autonomous Motorsport* team to tackle the competition, including the software architecture, simulation technology, and development workflow applied. The aim of the paper is to explain the relations and challenges behind certain design choices within the software stack and the respective outcomes during real-world testing.

## 1.2 | Vehicle platform

The official race vehicle of the IAC is the Dallara AV-21. It is based on the Dallara IL-15, which is used in the Indy Lights Series. It is equipped with a drive train consisting of a 2.0-L single-turbocharged engine and a 6-speed sequential semiautomatic gearbox. The retrofitting of the cars was mainly focused on the autonomous driving capabilities. Therefore, the basic parts of the drive train and the aerodynamic setup only received minor changes for the IAC, resulting in a similar behavior compared with the Indy Lights Series. It should be mentioned that the hardware platform comprising the conventional and the automated driving parts are equal for all teams. As a result, the performance of each vehicle in the competition solely relies on the implemented functional autonomy software of the teams. The autonomous driving parts are mounted in the driver's cockpit. The perception sensors and the computing platform replace the driver's seat, the actuation system is positioned in the footwell of the cockpit. The installed components are listed in Table 1 and are briefly described in the following.

The throttle, brake, and steering actuation is realized by a full Drive-by-Wire-system (DBW) by Schaeffler Paravan called *SpaceDrive II*. This embedded system consists of an electronic control unit

(a) (b)

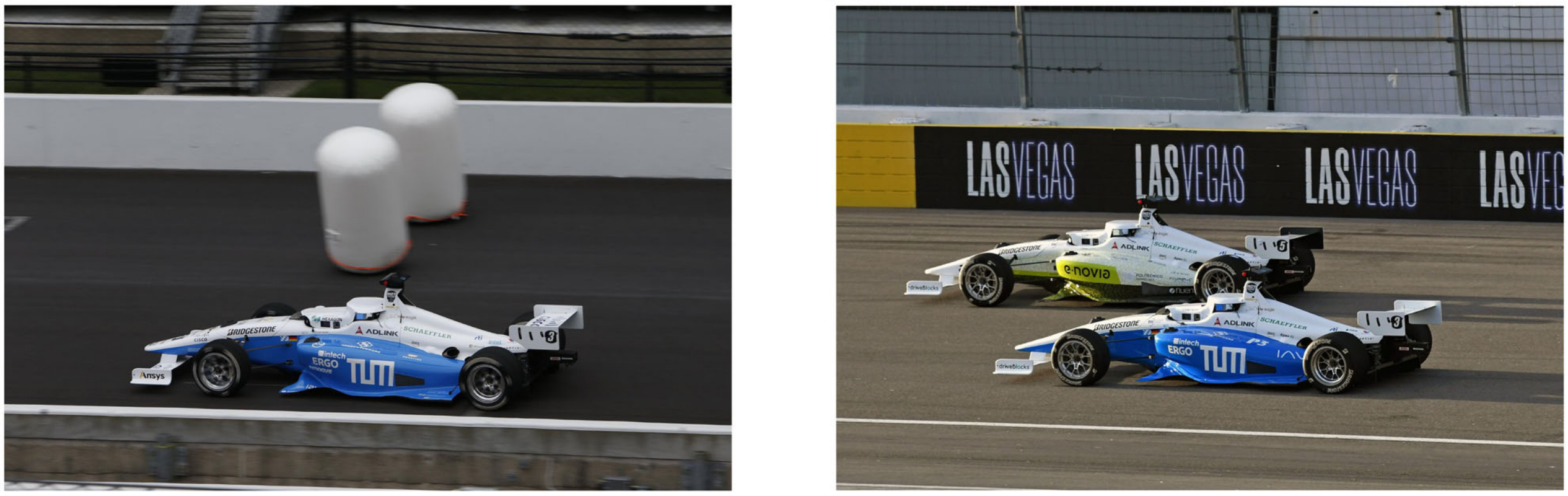

**FIGURE 1** The *TUM Autonomous Motorsport* racing software driving the AV-21 on different race tracks (Indy Autonomous Challenge, 2021). (a) Dallara AV-21 at Indianapolis Motor Speedway and (b) Dallara AV-21 at Las Vegas Motor Speedway. [Color figure can be viewed at wileyonlinelibrary.com]

**TABLE 1** Overview of the automated driving parts for the AV-21

| Component | Manufacturer | Model |
|---|---|---|
| DBW system | Schaeffler Paravan | SpaceDrive II |
| DBW interface | New Eagle | GCM 196 Raptor |
| ECU | Motec | M142 |
| Power management | Cosworth | IPS-32 |
| Computing platform | ADLink | AVA-3501 |
| Network switch | Cisco | IE500 |
| GNSS receiver | Novatel | Pwrpak 7d Receiver |
| LiDAR | Luminar | H3 |
| Camera | Allied Vision | Mako G319C |
| Side RADAR | Aptiv | MRR |
| Front RADAR | Aptiv | ESR 2.5 |

Abbreviations: DBW, Drive-by-Wire-system; ECU, electronic control unit; GNSS, Global Navigation Satellite Systems; LiDAR, Light Detection and Ranging; RADAR, radio detection and ranging.

and servo motors to receive braking and steering signals from the software and to execute them fulfilling real-time constraints. The overlying interface is realized by a New Eagle GCM 196 Raptor unit, which also handles the communication to the control unit of the combustion engine and the low-voltage power management. The communication on the actuation side is realized via the serial bus system Controller Area Network (CAN).

The core part of the automated driving hardware is an ×64-based computing platform. It is an ADLink AVA-3501, a modified version of the DLAP-8000. It comes with the 8-core Intel Xeon E-2278GE central processing unit (CPU) with 64 GB RAM and the Nvidia Quadro RTX8000 graphics processing unit (GPU) with 48 GB memory. Alongside the computing platform, a network switch establishes the connection to the sensors for perception and localization via Ethernet. The Global Navigation Satellite Systems (GNSS) is realized by two dual-antenna setups using Novatel Pwrpak 7d receivers. The perception sensor system consists of camera, radio detection and ranging (RADAR), and Light Detection and Ranging (LiDAR) sensors. In total there are six cameras installed, which are positioned to enable a full surround view. Similarly, the three LiDAR sensors are orientated in alignment to the vehicle heading and rotated around a vertical axis to ±120° such that LiDAR setup also covers in total 360°-field-of-view (FoV). The RADAR sensors are placed at the front and on both sides at ±90°.

## 1.3 | Related work

Teams with autonomous vehicles have already competed against each other in the past. The DARPA Grand Challenge (DARPA Grand Challenge, 2005; Buehler et al., 2007) was the first-long distance competition for autonomous vehicles. Participating university teams needed to build their own vehicle and write respective software capable of driving the car autonomously. The goal was to drive a predefined route of over 200 km fully autonomously without human interaction, and therefore, the vehicles needed to localize themselves, detect objects, and plan their path entirely on their own. As a successor, in 2007 the DARPA Urban Challenge (Buehler et al., 2009) presented a similar competition setup but now in an urban scenario. Furthermore, the cars needed to obey traffic rules, negotiate with other traffic participants to merge correctly, for example, into lanes and finish the race within 6 h.

Since these events, autonomous driving has become more and more relevant to the industry. New companies, like, Waymo, Zoox, and Cruise were established to develop a fully self-driving vehicle that operates the car in our transportation systems. At the same time, researchers began to use high-performance sports and race cars for their research purposes. This is because driving autonomously on the race track creates a variety of challenges for the autonomous software: localization and object detection at high speeds, trajectory and behavior planning in an adversarial environment, and control of the car at the dynamic limits of handling (Betz et al., 2018). The research in this field is mainly divided into soft- and hardware efforts.

### 1.3.1 | Software

A race track typically consists of a single lane as a driveable area with inner and outer bounds that are defined by curbs and none-driveable areas, like, grass and gravel. In addition, walls consisting of tires or stone surround the track to keep the car inside the race track in case of an accident. In the field of perception, researchers use the unique environment of the race track to demonstrate large-scale mapping with fewer features (Nobis et al., 2019) as well as localization at high speeds (Renzler et al., 2020; Schratter et al., 2021). Since the FSD competition requires the teams to drive and localize at the same time, the teams present Graph-SLAM (Andresen et al., 2020; Large et al., 2021) and Recurrent Neural Network-based methods (Srinivasan et al., 2020) for localization and state estimation of the FSD vehicle. In addition, the FSD competition provides yellow and blue cones as the race track and the teams need to detect those cones at high vehicle speeds. As a result, particular applications of YOLO-based methods are used to detect the cones (Dhall et al., 2019; Strobel et al., 2020).

In the field of path planning, authors focus on global, local, and behavioral planning. The global planning algorithms provide an optimal racing line for the whole race track. This racing line is the fastest trajectory for the vehicle that needs to be followed when there are no opponents around the car. Under specific optimization objectives, like, minimum curvature (Braghin et al., 2008; Heilmeier et al., 2019), minimum time (Christ et al., 2019; Pagot et al., 2020), and minimum energy (Herrmann et al., 2019) there are a variety of solutions to this problem. Local planning aims to achieve a high planning horizon for recursive feasibility while avoiding opponent vehicles with evasive maneuvers at high speeds. There are three main approaches for planning a local trajectory on the race track. First, the global plan can be adjusted and modified via an additional optimization (Kapania et al., 2016; Subosits & Gerdes, 2019). Second, multiple dynamically feasible trajectories are sampled. On the basis of racing-specific cost functions, the best trajectory that avoids obstacles is selected (Liniger et al., 2014; O'Kelly, Zheng, Jain, et al., 2020). Third, sampling-based methods provide an efficient but nonoptimal technique to randomly sample the free space around obstacles and find a possible trajectory (Arslan et al., 2017; Feraco et al., 2020). Finally, the work in the field of behavioral planning covers the task of planning the behavior of the car under high uncertainty and defining interactions with noncooperative agents. This type of behavioral planning for race cars is done either by designing multiple cost functions with weighting and then selecting the trajectory with the lowest overall cost (Liniger & Lygeros, 2015; Sinha et al., 2020) or by combining the local planner with game-theory methods (Notomista et al., 2020; M. Wang et al., 2021). Especially the ladder one showed the possibility of advanced cutting and blocking maneuvers (Liniger & Lygeros, 2020) which is crucial for the race car to succeed on the race track.

Finally, in the field of control, the goal is to handle the vehicle at the limits and track a reference trajectory as accurately as possible: low lateral tracking errors, low heading tracking errors, and low-velocity tracking errors. Another goal is to achieve high control frequencies with the available computation hardware for real-time high-speed driving. Research in this field uses an enhancement of classical control approaches to maximize the lateral and longitudinal tire forces (Fu et al., 2018; Kapania & Gerdes, 2015). A big part of the research applies Model Predictive Control (MPC) methods in some variations (Gandhi et al., 2021; Verschueren et al., 2016). The MPC solves a finite-time optimal control problem (OCP) and computes an optimal sequence of vehicle state and control inputs (steering and acceleration) based on a specific vehicle dynamics model (kinematic, linear single-track, and nonlinear single-track model). Lastly, since the autonomous race car is driving repeatably around the track for multiple laps, it is suitable for the application of *Iterative Learning Control* methods. With these data-driven approaches, algorithms are displayed that learn the control gap over time and apply afterward, for example, corrective steering input, to achieve a faster lap time (Hewing et al., 2018; Rosolia et al., 2017).

Furthermore, in addition to this classical perception–planning–control work, many researchers are focusing on full or partial end-to-end approaches that leverage the usage of deep neural networks (DNNs) or reinforcement learning (RL) methods. The racing task provides a clear objective function (fastest lap time) for the algorithm training and the race track provides with its clear driveable area and one class of objects a perfect proving ground. Researchers in this field displayed partial end-to-end approaches (Lee et al., 2019; Weiss & Behl, 2020) that combine DNNs with MPC methods to create and follow dynamic trajectories. In addition, by using algorithms from the field of RL (e.g., Soft-Actor-Critic and Q-Learning), researchers were able to demonstrate how to train an agent to drive fast (de Bruin et al., 2018; Jaritz et al., 2018), how to train an agent to overtake other agents on the race track (Song et al., 2021) and how to bridge the sim-to-real gap with model-based RL approaches (Brunnbauer et al., 2021).

### 1.3.2 | Hardware

Besides pure software development efforts, in the last years, various hardware platforms for the purpose of autonomous racing have been displayed. First, small-scale vehicles based on remote-controlled cars are used to test newly developed algorithms quickly. Those vehicles are equipped with sensors (Camera, LiDAR, and Inertial Measurement Unit [IMU]) and computation hardware to run the autonomous driving software. Researchers display hardware in a 1:43 scale (Liniger et al., 2014), 1:10 scale (Balaji et al., 2020; O'Kelly, Zheng, Karthik, et al., 2020), and 1:5 scale (Goldfain et al., 2019). The FSD competition covers a large part of the field of autonomous small-scale racing. Here university teams build their own 1:1.5 racing vehicles (Zeillinger et al., 2017) that need to drive autonomously around the race track in various competitions. The teams use these vehicles afterward for additional research and display both full autonomous driving stacks (Kabzan et al., 2020; Nekkah et al., 2020; Tian et al., 2018) as well as individual algorithm developments (Andresen et al., 2020; Large et al., 2021).

Full-scale vehicles are also used for autonomous racing research, apart from these small-scale race cars. In particular, these vehicles are high-performance sports cars that are used for autonomous handling at the limits (Funke et al., 2012; Theodosis & Gerdes, 2012) or autonomous drifting with high side-slip angle (Goh et al., 2019; Hindiyeh & Gerdes, 2014). In 2017, the company *Roborace* designed a special autonomous race car based on a LeMans-Prototype chassis. This vehicle was equipped with sensors, actuation, and computation hardware to drive autonomously around the race track. *Roborace* gave interested student teams the opportunity to use this race car which displayed research in the field of localization (Massa et al., 2020; Zubaca et al., 2020), high dynamic path planning (Caporale et al., 2018; Stahl et al., 2019b), software development (Betz et al., 2019; Hermansdorfer et al., 2020), and control (Buyval et al., 2017; Wischnewski, Betz, et al., 2019). In addition, *Roborace* organized different competitions called *Season Alpha* and *Season Beta* that consisted of single- and multivehicle events on various race tracks. Finally, the IAC vehicle is the latest autonomous race car that was designed for research and competition purposes and is further explained in Section 1.2.

In summary, we can say that autonomous racing is an emerging topic in the field of robotics and intelligent vehicles (Betz et al., 2022). With the rising number of active researchers in this area providing both software and hardware developments, the community is constantly growing. With the setup of the IAC we see the first-time vehicles competing against each other at high speeds and high accelerations—entirely autonomously.

## 1.4 | Contributions and outline of the paper

In this paper, we present the efforts in the software development of the *TUM Autonomous Motorsport* team for participating in the IAC. This work builds upon (Wischnewski et al., 2022) and has four main contributions:

1. We provide a holistic view of the software architecture and design decisions made during the development of the *TUM Autonomous Motorsport* software stack for high-speed autonomous racing.
2. We elaborate on the development and testing workflow and their impact on the achievements made during the both IAC competitions.
3. We provide an evaluation of all developed software modules in a full software stack. The results obtained in this full-stack evaluation include implications that might be hard to find in isolated studies and research projects.
4. Finally, we share experimental results of single-vehicle as well as two-vehicle racing scenarios with speeds of up to 270 $km\,h^{-1}$ and accelerations up to 28 $ms^{-2}$.

The paper is structured as follows: Section 2 introduces the software architecture and gives insights into the applied algorithms and concepts. Section 4 describes the event formats as well as the results and findings during these experiments. Finally, Section 5 summarizes the learnings and conclusions from this project. It outlines streams of future work and potential areas of technology transfer from the race track to series production vehicles.

# 2 | TUM AUTONOMOUS MOTORSPORT SOFTWARE

This section deals with the software that the *TUM Autonomous Motorsport* team developed for participation in the IAC. After the introduction of the overall software architecture in Section 2.1 the specific software modules are presented in the order of application in the overall stack. Additionally, related topics such as middleware and software latency as well as our development infrastructure consisting of Software-in-the-Loop (SiL) and Hardware-in-the-Loop (HiL) simulation are displayed.

## 2.1 | Architecture

The software architecture (Figure 2) employs a classical separation into three main areas: perception, planning, and control. The perception module leverages RADAR, camera, and LiDAR to detect opponent vehicles. The LiDAR detection is done with two different strategies to increase the reliability: First, a deep learning-based approach is utilized to classify race vehicles in the point-cloud data. This algorithm is specifically trained on race vehicles and shows high detection performance. At the same time, it is prone to overfitting and will not detect other classes of objects which might appear on the track due to unforeseen circumstances. The second approach aims to overcome these deficiencies with a geometric clustering approach. Even though it takes into account basic geometric information about the considered objects, it is capable of detecting arbitrary classes of objects. These two main pipelines are accompanied by a pipeline for camera-based detection. It uses a bounding box

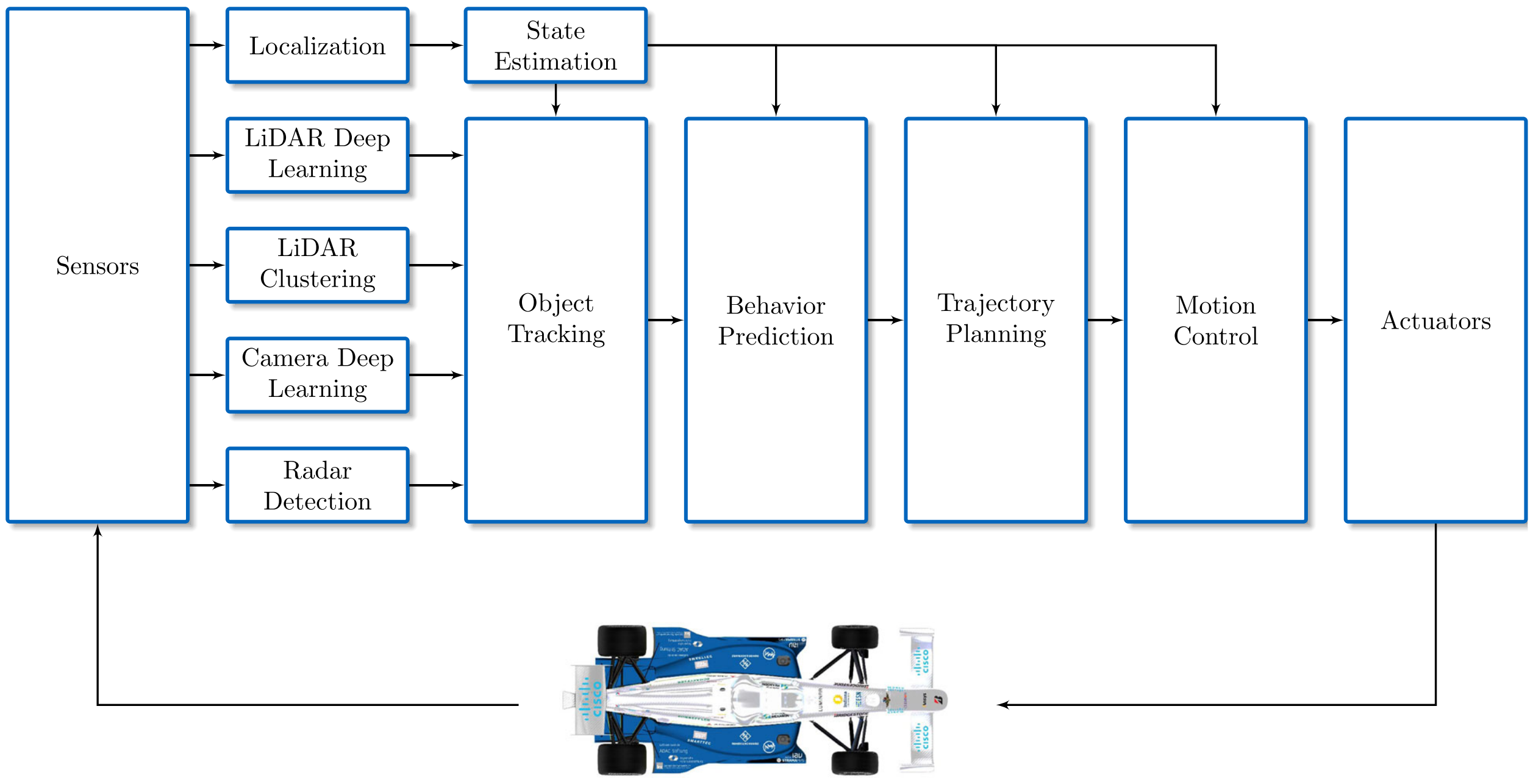

**FIGURE 2** Software architecture of the *TUM Autonomous Motorsport* team. LiDAR, Light Detection and Ranging; TUM, Technical University of Munich. [Color figure can be viewed at wileyonlinelibrary.com]

approach in conjunction with a known-height assumption for distance estimation. While this enables large detection distances, the transformation errors resulting from track banking and different vehicle orientations lead to greater positional uncertainties than the LiDAR pipelines. Finally, RADAR detection completes the set of detection algorithms. Its main strength is the ability to measure the velocity difference between an object and the ego-vehicle. This improves the transient performance of opponent velocity estimation, which is key for reliable driving performance in highly dynamic scenarios with limited sensor range.

The detected objects are fed into an object tracking algorithm, which serves two purposes: The matching of already tracked objects and incoming measurements as well as the temporal fusion of the detections. The first task is executed by an algorithm based on the Hungarian assignment method (Kuhn, 1955). The second task is achieved via a Kalman-Filter-based approach with a constant velocity and turn rate dynamic model. This filter creates a position history of the fused observations for each identified vehicle on the track. This history serves as a basis for the prediction of future behavior, which has been implemented in two different ways: First, a physics- and track-informed prediction; second a data-driven approach. The resulting predictions are the most likely outcomes for the behavior of the opponent vehicles and are handed over to the trajectory planning module.

The trajectory planning module is based on a combination of a sampling and a graph-search strategy. This makes it possible to resolve the combinatorial nature of the multivehicle planning problem efficiently. The planned trajectory is approximately 5 s long and has to be updated frequently to take into account new information about the behavior of other race participants. The target path and target velocity profile are handed over to the motion control module which utilizes an MPC algorithm to attenuate disturbances and optimize the coarse output of the trajectory planning to generate smooth vehicle behavior. This is enabled via the introduction of a safe driving tube which is assumed to be collision-free for approximately two times the vehicle width. Finally, the state estimation provides a consistent estimate of the vehicle position and motion state to all software modules. It leverages multiple localization sources (two global positioning system [GPS] and one LiDAR localization) as well as two IMUs to achieve reliable operation and fault tolerance.

## 2.2 | Localization

High precision and low latency localization is a key challenge of autonomous racing (Lingemann et al., 2005). The sensors used for this application are two Novatel GNSS receivers with integrated IMU and two antennas each. One has the two antennas at the left and right sidebox and the other has them at the nose and on top of the main roll hoop.

In Sauerbeck et al. (2022), we introduced a localization algorithm that uses camera images and LiDAR point clouds for ego pose estimation. However, real-world testing showed no benefit over a redundant differential GNSS setup at open-sky race tracks. The final localization and state estimation were mainly based on a fusion of the two GNSS signals and their IMU units. Therefore, an enhanced version of the Kalman-Filter approach presented in Wischnewski, Stahl, et al. (2019) is used. This approach is based on a two-dimensional (2D) point mass model to represent the vehicle dynamics. Since detailed data of the vehicle setup and the used tires were not available, this approach can outperform approaches with a more detailed vehicle model (Wischnewski, Stahl, et al., 2019). The measurement quality of the two GNSS receivers was determined empirically and the weighting of the sensors was specified accordingly. Since the differential heading calculated from the GNSS receiver by antenna positions does not exhibit reliable behavior, we use a heading estimation based on the derivation of velocity, which provides small errors when the vehicle turns. To account for the track banking (up to 9° in Indianapolis and 20° in Las Vegas), additional banking information is used. The used map consists of 2D track boundaries and a one-dimensional (1D) banking map along the race track. Lateral differences in banking can be disregarded because measurements proved them to be small enough. Moreover, this avoids numerous exploitation runs. The banking information allows the compensation of the banking in the measured accelerations and calculate the Kalman Filter as on a plane. As shown in Equations (1) and (2), only the lateral acceleration is compensated. $a_x'$ and $a_y'$ denote the accelerations used for the state estimation. $a_{x,\text{meas}}$ and $a_{y,\text{meas}}$ are the measured accelerations received from the IMU. $\theta(s)$ is the banking angle at the corresponding longitudinal track coordinate $s$, and $g$ is the gravitational constant.

$$a_x' = a_{x,\text{meas}}, \tag{1}$$

$$a_y' = a_{y,\text{meas}} \cdot [\cos(\theta(s)) + \tan(\theta(s)) \cdot \sin(\theta(s))] + g \cdot \tan(\theta(s)). \tag{2}$$

The 2D track boundaries were generated with laser scans and known ego-position. To obtain the 1D banking map, the residuals from the state estimation Kalman Filter were utilized. The Kalman Filter for localization and state estimation is implemented in *Matlab Simulink* and deployed to the car via C-code generation. It is executed in the same process as the vehicle controller at a frequency of 100 Hz.

## 2.3 | LiDAR preprocessing

The first step in the LiDAR object detection pipeline is the preprocessing to reduce the number of points captured by the three LiDAR sensors. Each unit sends raw point cloud to the LiDAR sensor driver, where the point clouds are directly fused and transformed into the vehicle's coordinate system. The output of the driver is a raw point cloud covering a horizontal FoV of 360° and a vertical FoV of 17.5° and 20° for the sections of the front and left/right LiDAR sensors, respectively. The front LiDAR vertical FoV is lower compared with the left/right LiDAR sensors' vertical FoV to achieve a higher resolution, which can be beneficial with distant objects. With a sensing frequency of 20 Hz, the LiDAR sensors have 32 vertical layers. The distribution of these layers can be changed at runtime. We use this feature to dynamically adapt the high-density layer region to the region of interest (ROI). Due to the track's banking, the vertical ROI has different positions based on the location of the vehicle on the track. On the track's straights, the ROI is centered in front of the vehicle, whereas in the banked turns, the ROI is shifted to the top.

The point cloud serves as input for both LiDAR detection algorithms. Their main task is the detection of objects on the track. These objects usually consist of only a few points in the point cloud, and the number of points decreases significantly with increasing distance between the vehicle and the object. As a result, only a few points are relevant for the driving task and the rest of the points within a point cloud should be filtered before it is passed to the object detection algorithms. This not only increases the algorithm's performance, but also reduces the computational load, the data transfer times and lowers the overall latency. Hence, the vehicle can react faster to the opponent's changes in position and orientation. However, this comes at the cost of additional computational load and calculation time for the preprocessing itself. Therefore it is necessary to employ lightweight and efficient preprocessing algorithms. Since the number of points per object is low, especially at distances beyond 50 m, the preprocessing algorithms should not reduce the point density of the relevant objects.

Given these initial constraints, we develop a point cloud preprocessing pipeline consisting of three sequential algorithms: conditional removal, voxel downsampling, and ground filter (Figure 3). Each algorithm is described briefly in the following. The order of the three algorithms is based on the algorithm's ability to handle large point clouds with low computational load and time. The ground filter benefits from a lower input point number as opposed to conditional removal, which can handle an arbitrary number of points without additional computation time.

*Conditional Removal* is a method to extract the relevant ROI from a point cloud. The goal is to remove any points which are outside of the race track, such as reflections from buildings or the stands, based on geometric filtering. Points that meet certain criteria are labeled as not relevant and are therefore removed from the point cloud. Conditional removal is performed in local vehicle coordinates and no information on the vehicle's global position is used. Hence, conditional removal is based on the assumption that the vehicle's heading is roughly parallel to the direction of the racing line.

*Voxel Downsampling* is a method to compress the information about multiple points within a certain area into a single point. The entire point cloud is divided into a grid with a fixed voxel size by using an algorithm from the Point Cloud Library (Rusu & Cousins, 2011). We use cuboid voxels of different sizes for each cuboid side. The selection of the length, height, and width of the voxels is based on the expected point cloud shape of the relevant objects, which are mainly race vehicles. A voxel size of 0.15 m/0.1 m/0.05 m for $x/y/z$ reduces the number of points in close range but keeps the point cloud structure for objects at higher distances. Beyond a threshold of 150 m, the points are not voxelized. For each voxel, the average in each $x/y/z$ of all points within the voxel generates a new output point representing this voxel. In case no points are found within the voxel, an output point is not created for this voxel. The resulting output point cloud resembles the input point cloud with fewer points.

The *Ground Filter* uses a ground segmentation algorithm to detect points belonging to the ground and filters these out of the point cloud. The usage of a neural network trained in a supervised fashion for this task is not an option due to the lack of a data set. Data from a race track including banking with pointwise labels are not available and are not feasible to create due to limited testing time. Therefore, we employ a ray-ground filter, based on the implementation of Autoware.Auto (The Autoware Foundation, 2021), which follows the ideas of Cho et al. (2014).

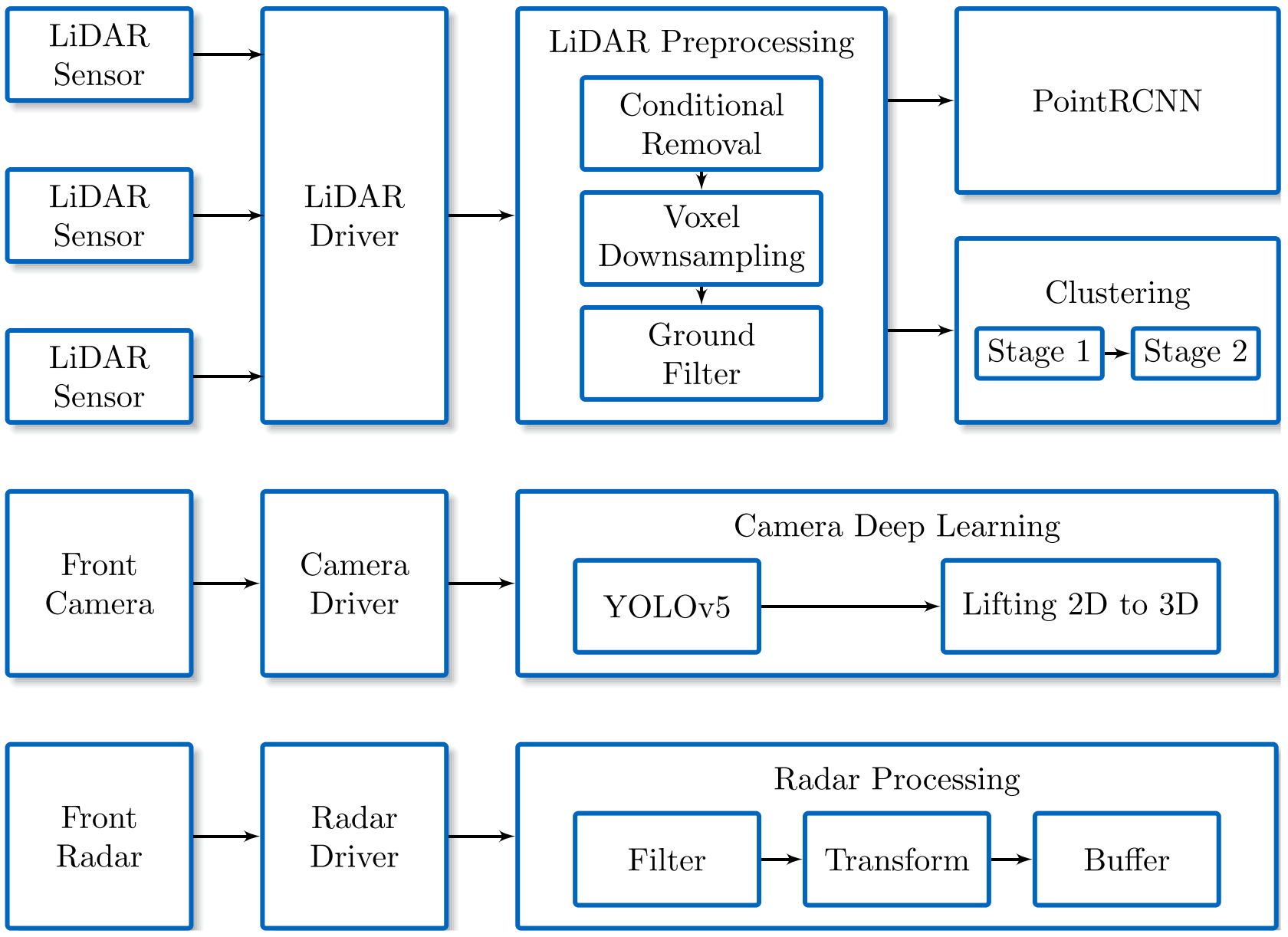

**FIGURE 3** Overview of the object detection pipelines. 2D, two-dimensional; LiDAR, Light Detection and Ranging; RCNN, Regions-based Convolutional Neural Network. [Color figure can be viewed at wileyonlinelibrary.com]

We test a point cloud distortion correction algorithm based on pointwise time stamps to remove the distortion originating from the difference in capture time for all points within a single-point cloud. The effect of distortion can be seen especially at higher speeds. For example, at a vehicle speed of $60\,m\,s^{-1}$ and a LiDAR refresh rate of 20 Hz, the vehicle moves 3 m between the first and last captured points in a single-point cloud. Since the relevant objects travel with roughly the same speed as the ego-vehicle and the relative speed difference—and therefore the distortion of these objects—is low, the effect can be neglected. Hence, we do not actively use distortion correction during the race.

The performance of the individual LiDAR preprocessing algorithms is depicted in Figure 4. Outliers with a lower number of points, especially at the raw input point clouds, emerge when only one or two of the three LiDAR sensors send data, which occurs occasionally for single time steps. Overall, the preprocessing pipeline reduces the point cloud size by more than 80% with a total calculation time of around 22 ms, including data transfer between the algorithms. Figure 5 illustrates the output of the three preprocessing algorithms on an exemplary point cloud. Although the visual difference between (b) and (c) is hard to identify, the voxel downsampling step halves the number of points. The reduced point clouds retain the relevant information of each raw point cloud and serve as input for the following object detection algorithms.

## 2.4 | Object detection—LiDAR deep learning

For the detection of opponent race vehicles, we employ a neural network that uses preprocessed point clouds (Section 2.3) as input. Specifically, we select the two-stage PointRCNN (Shi et al., 2019), which ranked at the top of the KITTI Benchmark at the time of selection (Geiger et al., 2012). To fit our needs for detecting race vehicles, we modify this network as described in the following. First, we adapt the network configuration to enable a 360° horizontal FoV. Additionally, we move the reference system of the detections from the front camera (default KITTI Benchmark) to the vehicle rear axle. Since there is only one type of race vehicle to detect, the network only has to predict one class. Finally, we manually fine-tune parameters, such as detection thresholds, for the best performance in our use case.

The network is trained in a supervised fashion. Labeled point cloud data sets with race vehicles were not available until our first tests on the race track. Therefore, we use initial training data

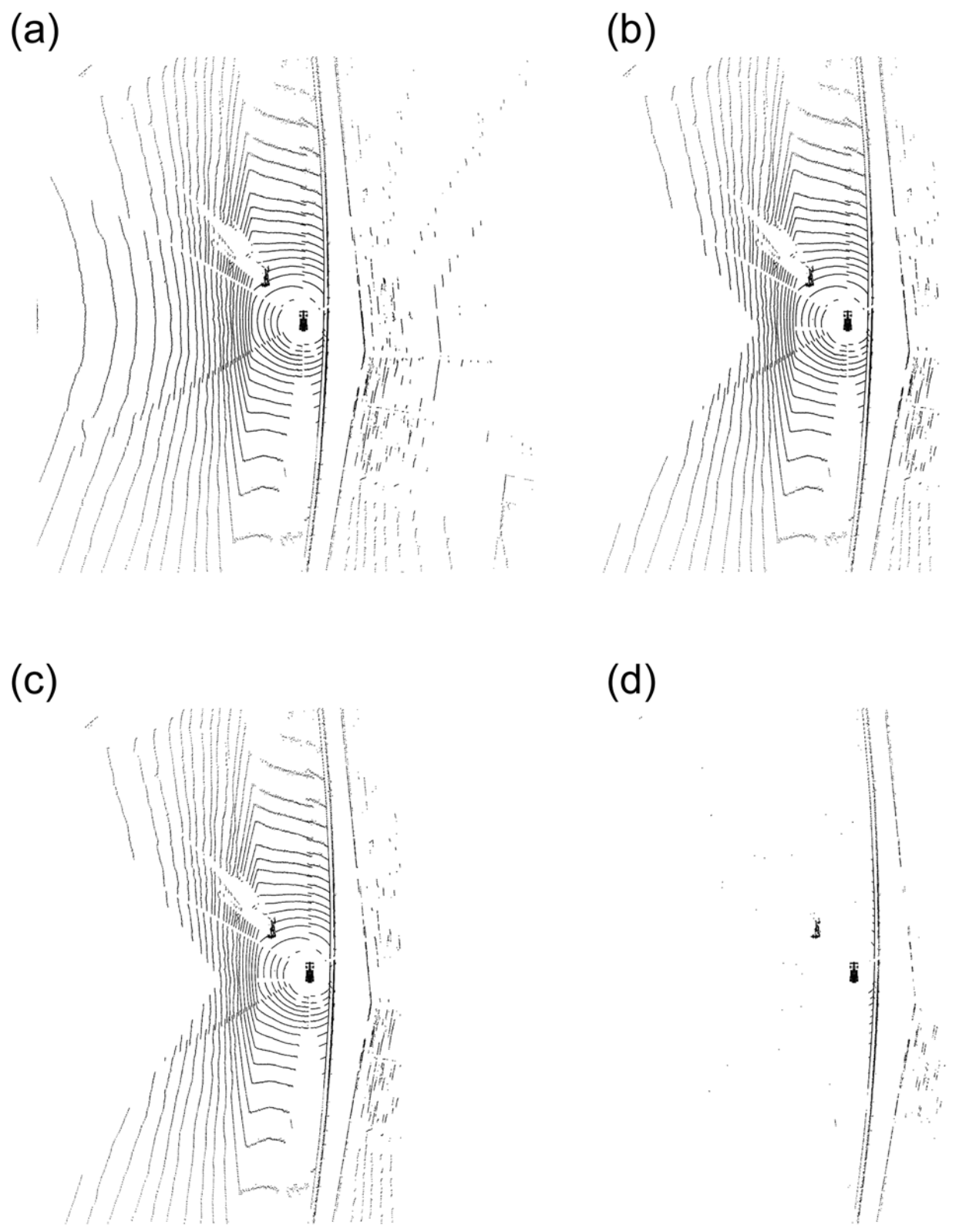

**FIGURE 5** Visualization of the three LiDAR preprocessing steps on an exemplary point cloud. (a) Raw input 74,302 points, (b) after conditional removal 66,585 points, (c) after voxel downsampling 29,502 points, and (d) after ground filter 11,995 points. LiDAR, Light Detection and Ranging.

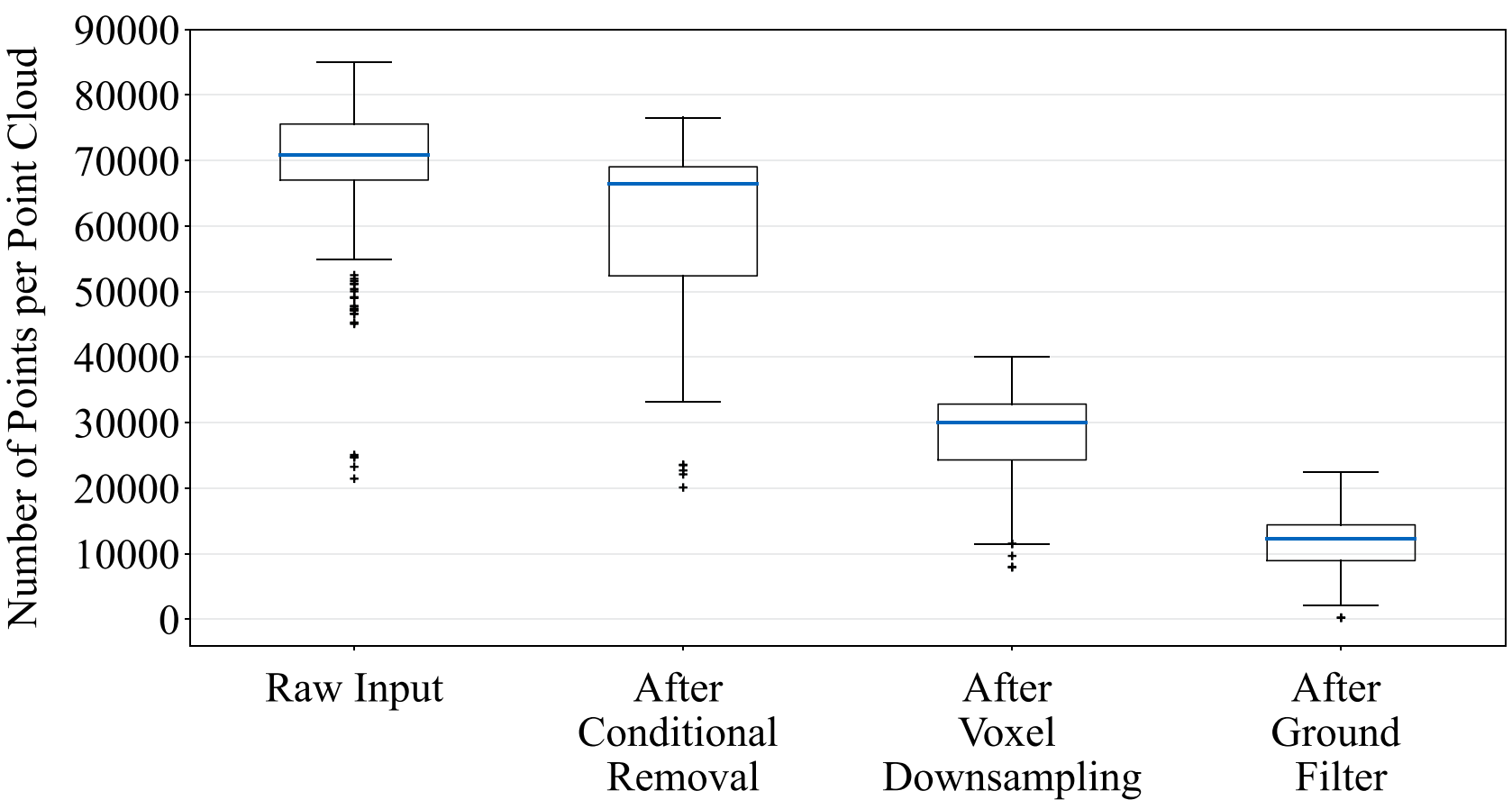

**FIGURE 4** Breakdown of the point reduction for each LiDAR preprocessing algorithm, based on 32,568 point clouds captured during the final run of the Autonomous Challenge at CES on January 7, 2022. LiDAR, Light Detection and Ranging. [Color figure can be viewed at wileyonlinelibrary.com]

generated in our simulator (Section 3.2). Additionally, we manually label data recorded from the first real-world multivehicle test sessions. Once the network's performance reaches a satisfactory state, that is, it detects the race car in every other point cloud, we use it to autolabel the recorded data and supervise the results. Both training and inference are conducted on a GPU to leverage the benefits of parallel processing. Deployed on the race vehicle, the network runs at a frequency of around 12 Hz.

## 2.5 | Object detection—LiDAR clustering

The LiDAR object detection neural network can detect only objects it has encountered during training. Since we generate a data set containing race vehicles only, other unstructured objects like debris cannot be detected by the neural network. Therefore, we employ a second object detection algorithm that can detect any kind of object on the race track. Furthermore, this competitive pipeline complements detections from the neural network. This increases redundancy in case either of the algorithms misses an object. In case both algorithms detect the same object, even with different extends, the output is fused in the object fusion and tracking algorithm (Section 2.8). The pipeline is based on a classical machine-learning algorithm. In detail, we employ a grid-based Euclidean-cluster-extraction algorithm, which operates in two stages.

The first clustering stage is specialized to detect small clusters within the preprocessed point cloud. Small clusters usually include parts of a race vehicle, such as the front wing or wheels. Clusters larger than the dimensions of a race vehicle are disregarded. In the second clustering stage, the remaining clusters are combined into larger clusters, ideally returning one cluster per race vehicle or object. This stage is also followed by a threshold step, in which clusters larger than race vehicles are not considered. This threshold step works only as long as there is only one opponent at a time on the race track, which was always the case during testing and the events. Three-dimensional (3D) bounding boxes are calculated based on the minimum and maximum extends in $x/y/z$ of the points in each cluster. These boxes are oriented in the same direction as the ego race vehicle. The clustering algorithm runs with a frequency of 20 Hz on one core of the CPU.

## 2.6 | Object detection—camera deep learning

Object detection using cameras provides additional redundancy. However, the projection of the 3D world onto 2D images entails a loss of information and direct detection in 3D space is not possible. Our approach to solving this challenge is the detection in the 2D space and subsequent recovery of the 3D information based on a priori knowledge. In detail, we use YOLOv5 (Jocher et al., 2021) for object detection on 2D images. The input is the image recorded by the front camera and the output are 2D bounding boxes of all detected objects on the input image. The recovery of the object's 3D information, that is, the relative $x$ and $y$ distance to the ego-vehicle, is based on a pinhole model of the camera. Using the intercept theorem, the distance $d_{obj}$ of the object to the camera is a function of the camera focal length $f$, the 2D bounding box height pixel count $n$ (assuming a known pixel size), and the real race vehicle height $h$

$$d_{obj} = \frac{fh}{n}. \tag{3}$$

We calculate the rotation of the object around its vertical axis using the ratio of the bounding box width and height and compare this to the known width-to-height ratio of the real race vehicle. The resulting rotation angle estimate is not unique, that is, the object can be rotated to the left or right with the same angle. On the basis of the position of the oval track, one of the two solutions is more likely. We also experimented with predicting the rotation angle directly using YOLOv5 with an additional output per predicted bounding box, but the results were inferior to the rotation estimation from the width and height ratio.

## 2.7 | Object detection—RADAR

The RADAR detection pipeline (Figure 3) extends the set of perception algorithms with an additional and independent object detection method to further increase the functional safety of the vehicle. The main benefit of this pipeline is the utilization of the RADAR sensor's ability to directly measure the relative velocity of the opponent vehicle via the Doppler effect. In addition, the RADAR sensor represents the sensor with the highest detection range on the straights and therefore enables early object tracking with an accurate speed estimation. However, the RADAR sensor is limited to a maximum of 64 detections per cycle and is subject to a high number of false positives. Therefore, a dedicated RADAR processing pipeline had to be developed to deal with these challenges.

The main purpose of the RADAR processing pipeline is the filtering of the input data to extract the objects of interest from the surroundings. The applied filter separates the incoming objects, based on their absolute velocity, to isolate the dynamic objects from the static environment. To achieve this, a threshold-based filter is used and tuned for a racing application. Finally, the filtered objects are transformed to the vehicle frame and stored within an object buffer to supply the downstream modules with a fixed frequency of 20 Hz.

## 2.8 | Object fusion and tracking

This section outlines the software module of object fusion and tracking. For more detailed information about the fusion and tracking task, the reader is referred to Z. Wang et al. (2020). The object fusion handles multiple object lists that originate from different perception pipelines. Ultimately, this algorithm combines the given information to output a unified object list. As Figure 2 reveals, the perception pipelines work independently from each other and output individual

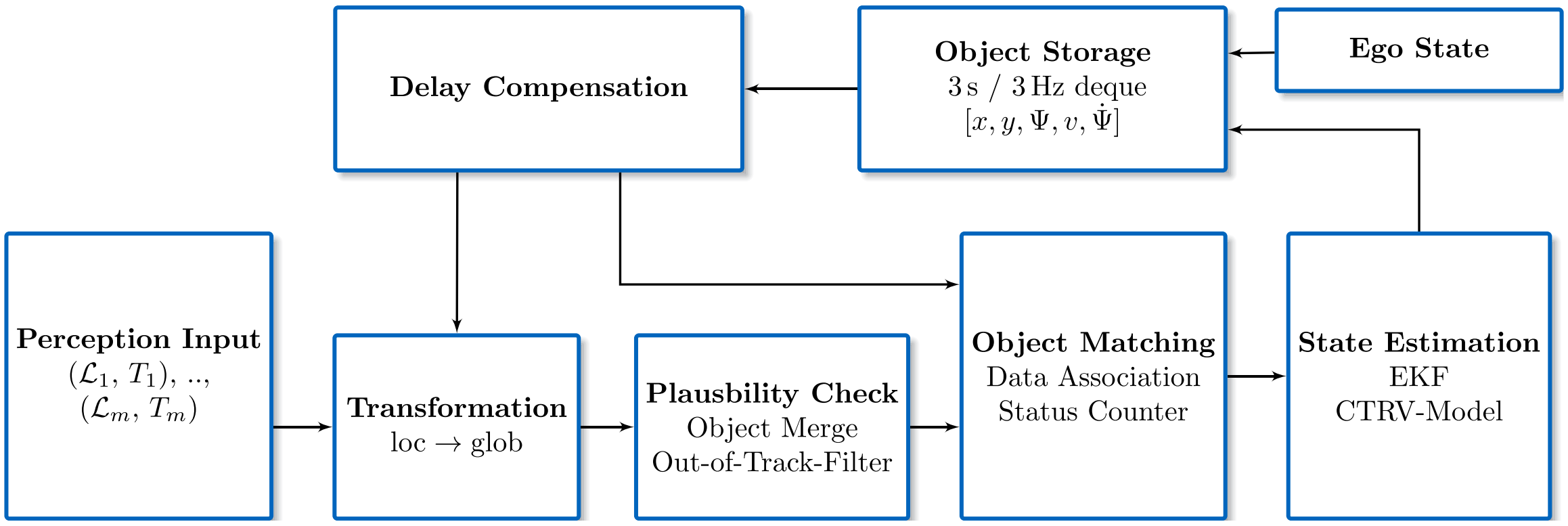

**FIGURE 6** Overview of object fusion and tracking. CTRV, Constant Turn Rate and Velocity; EKF, Extended Kalman Filter. [Color figure can be viewed at wileyonlinelibrary.com]

object lists. This late fusion approach allows us to incorporate a variable number of perception pipelines without any dependencies. This is especially beneficial to conduct real-world tests with basic perception pipelines in the absence of rich data to train and develop more comprehensive algorithms. The object tracking addresses the estimation of the detected objects' dynamic states, which is realized by the Extended Kalman Filter (EKF; Jazwinski, 1970) based on a Constant Turn Rate and Velocity (CTRV)-model. An important feature to realize this is the matching between previously estimated objects and new objects provided by one of the perception pipelines. Figure 6 outlines the module sequence, which is described in the following.

Input to the module is both the dynamic vehicle ego-state and the perception input. The latter one consists of an object list $\mathcal{L}_i$ and a sensor time stamp $T_i$ per perception pipeline. Our implementation is parametrized with $m = 4$ perception pipelines but is capable to scale up to an arbitrary number of pipelines. Depending on the applied algorithm and sensor the object states in the object list contain distinct variables. Due to the fact that the modules are not synchronized and the individual perception pipelines have different cycle times, the number of received objects lists varies. The perception input is processed sequentially, starting with the object list having the oldest sensor time stamp. The object list has to be transformed from the local vehicle coordinates to a global coordinate system, which is used for tracking, prediction, and trajectory planning. In this step, a yaw estimation based on the orientation of the track's center line is added in case the heading is not given by the respective perception pipeline. By this, the state estimation can be enhanced. The coarse assumption is handled by a high measurement uncertainty in the EKF update step.

Next, plausibility checks are conducted in two stages. In the first stage, multiple detections of a single object are merged which prevents multiple tracks of the same object. This is realized by kd-tree clustering (Maneewongvatana & Mount, 1999) and a fixed distance threshold for the cluster distance. In the second stage, we reduce the number of false positives with a map-based filter. This removes all objects which are outside the global track boundaries and is necessary due to the reflectivity of the pit and track wall.

The filtered object list is input to the object matching, which is based on the Hungarian Method (Kuhn, 1955). This combinatorial optimization algorithm solves the data association problem of the given old object list from the previous time step and the new object list. The applied cost function is the pairwise distance between $n$ old and $m$ new objects, which are assigned in an $n \times m$-matrix. With the constraint of a maximal valid matching distance, the solution of the assignment problem can result in the following cases:

- *New unmatched object*: There are more new objects than old objects ($n < m$) or matches are classified as invalid and the new object remains unmatched if the matching distance is above the threshold. The new unmatched object is assigned with a unique ID and a status counter is set up. Additionally, the CTRV-model in the state estimation module and object storage are initialized.
- *Old matched object*: There is a valid match between an old and a new object. In this case, the ID of the old object is assigned to the new one. The status counter is increased, an update step of the EKF is conducted and the resulting state correction is added to the object storage.
- *Old unmatched object*: There are more old objects than new objects ($n > m$) or matches are classified as invalid and the old object remains unmatched if the matching distance is above the threshold. The old unmatched object's status counter is decreased and the object storage is updated with the estimated state as no measurement update step is possible.

The idea of the status counter is to define the number of perception inputs without detecting a tracked object before it is discarded. A status counter is initialized with a positive integer for each new unmatched object. In case the object is matched by the next perception input (old matched object), the counter is increased by $X_{\mathrm{up},i}$. Otherwise, if the object is not matched (old unmatched object) the counter is decreased by $X_{\mathrm{low},i}$. If the status counter reaches 0, the object is removed from the storage and is not tracked anymore. The values for $X_{\mathrm{up},i}$ and $X_{\mathrm{low},i}$ are positive integers and depend on the perception pipeline $i$. The value of the status counter is limited by a maximal value $X_{max}$. By this, it is ensured that an object, which was successfully matched multiple times, is still removed quickly after it has not been detected anymore. The parameterization of the counter values depends on the sensitivity and specificity of the perception pipelines and the trade-off between recognizing objects preferably early when they enter the sensor range and discarding them if they are not detectable anymore.

The state estimation runs with a filter step size of 10 ms, which means that during each cyclic call of the module the forward integration of the EKF estimation step is executed multiple times. By this, the linearization error of the EKF remains within the tolerance. All estimation steps are stored in the object storage to ensure an equidistant sampling rate. However, the respective entries are replaced by the corrected position after a successful match and related EKF update. The CTRV-model is an appropriate choice for our use case of autonomous racing. On the one hand, the yaw rate is essential to accurately estimate the objects' motion in turns at high speed. On the other hand, the estimation of the acceleration is prone to oscillations. So we attempt to reflect this trade-off in the complexity of the model. The states of the tracked objects are stored in the object storage, which is a deque of 3 s with 100 Hz resolution. Each state comprises the 2D-position ($x$ and $y$), the orientation in the global coordinate system (heading $\Psi$), the speed $v$, and the yaw rate $\dot{\Psi}$.

For reliable object tracking at high speeds the consideration of delays resulting from the sensors and the perception algorithms is essential. The implemented delay compensation handles this task with a backward–forward iteration within the object storage during the object matching and the update step of the EKF. The backward iteration occurs before the object matching takes place. On the basis of the received sensor time stamp, the existence of objects in the object storage at the given time is checked. These historic object states including the ego-state are applied for the transformation and matching procedure. In case of a successful match, the state estimation is corrected at the given sensor time stamp and the corrected state is iteratively predicted up to the ego time stamp. New unmatched objects, which are initialized at the outdated perception time stamp are also iteratively predicted up to the ego time stamp. With this concept, the implemented delay compensation enables the synchronization of the tracked object states with the ego time stamp while still considering delayed perception inputs up to 200 ms and is one of the core features to enable high-speed multiobject racing.

## 2.9 | Prediction

There are some key aspects in which prediction in autonomous racing differs from that on public roads: First, the vehicles are in direct competition with each other. While all participants want to avoid a collision, they are not necessarily cooperative, but competitive. Second, there are no intersections, lanes, or traffic rules. While the absence of intersections or forks initially simplifies the prediction task, this task is complicated by the absence of traffic rules and lanes. This ultimately leads to situations where the lateral uncertainties of predictions take up the entire width of the race track because a wide variety of driving and behavior patterns are possible with the same initial situation. Third, multiple laps are run on the race track. So similar situations appear several times during a race and can be used as valuable information for future predictions. In the following, we will briefly present our approach to the trajectory prediction of adversarial vehicles, returning to these particular aspects. Current approaches in vehicle trajectory prediction are classified into three different categories (Karle et al., 2022): physics-based, pattern-based, and planning-based. Our approach incorporates components from all three of these categories. The prediction model is based on the following input data: a tracked object list (Section 2.8), an ego-state tracked over time (Section 2.12), and map information (Section 2.2). As output, a most likely trajectory for each vehicle on the track is sent to the planning module (Section 2.11) in the form of time-dependent $x-y$ positions over a time horizon of 5 s with a sampling rate of 5 Hz. We also incorporate uncertainties with a bivariate Gaussian distribution for every local point.

We build our approach on an long short-term memory (LSTM)-based encoder–decoder architecture. This network encodes the states of the predicted vehicle, as well as the track boundaries with LSTM layers. Unlike common approaches for road traffic (Messaoud et al., 2021), we avoid using an image-based map representation due to the simple track geometry, and instead make use of a much more efficient way by processing the left and right boundary similar to the ego-state history in a vector representation. The different input streams are fused by concatenation in the latent space. In contrast to previous work (Deo & Trivedi, 2018), we do not use LSTM decoders that generate corresponding time-dependent points as trajectories. The use of LSTM decoders has the definite disadvantage that the predicted trajectories are often physically infeasible and leave the track, for example. Hence, we extend this purely data-based approach to include physical knowledge: We identify basic trajectories that completely cover the output space of predictions by linear combination. Consequently, the neural network directly learns not the time-dependent positions, but weighting factors for the base trajectories; we call this mixture of data-based and physical approaches *MixNet*, which is discussed in further detail in Török et al. (2022). Furthermore, to make use of observations from past laps in similar situations we investigated an additional online-learning approach (Geisslinger et al., 2022), which adopts the weights in a neural network according to an observation loss. However, due to a lack of robustness (and the inability to recognize specific objects), eventually, we did not incorporate this online learning in the final software stack.

We also want to account for interactions on the race track in the sense that each vehicle's behavior does not depend only on its past states and the track boundaries, but also on the other vehicles around. This can be solved by learning interactions from a data set in the neural network, but the solution requires vast amounts of data from interactive scenarios. To account for the interaction of different vehicles (including the ego-vehicle) in our prediction network, we modify the predictions with a subsequent planning-based approach. For this purpose, we first predict each vehicle, including the ego-vehicle, independently. The predictions are checked for collisions and only if a predicted collision occurs are the trajectories modified. On the basis of the racing rules, which are similar to those of Formula (1), we make the simplifying assumption that the rear vehicle must react to the leading vehicle and adjust the prediction of the rear vehicle accordingly. To do this, we utilize fuzzy logic to decide whether an overtaking event will occur and whether it will occur on the left or right side.

We also use a priori quality measures of our MixNet predictions, which are described in Török et al. (2022). Once a quality measure exceeds a certain threshold we fall back to a simplified, naive prediction. In this naive prediction, we use a constant velocity profile originating from the object tracking (Section 2.8) and assume that the vehicle will hold its line in terms of lateral positioning between the left and right boundaries. This simple, but effective approach proved to be sufficient for the passing competition at the AC@CES (Section 4.2).

## 2.10 | Global planning

The global planning module builds upon the work of Christ et al. (2019). Their work describes an OCP to solve a minimum lap time problem for an autonomous race car, which is transcribed into a nonlinear program via direct orthogonal collocation. The OCP is formulated using the CasADi modeling language (Andersson et al., 2019), and subsequently solved using the Interior Point OPTimizer (Wächter & Biegler, 2006).

The optimization problem to be solved minimizes the achievable lap time $t_l$,

$$\min t_l = \int_0^{S_\Sigma} \frac{dt}{ds} ds = \int_0^{S_\Sigma} \frac{1 - n\kappa}{v \cos(\xi + \beta)} ds, \quad (4)$$

while simultaneously adhering to the constraints stemming from the driving dynamics of the vehicle, which we describe as a nonlinear double-track model. With this, we can ensure that the realistic vehicle dynamic behavior is captured—especially the nonlinear behavior of the tires. The traveled distance $s$ along the reference line is used as the independent variable. The race track geometry is described by the curvature profile $\kappa$, $v$ denotes the velocity on the racing line, $n$ the lateral distance to the reference line, $\beta$ the side-slip angle, and $\xi$ the relative angle of the vehicle to the tangent on the reference line. For further details on the formulation of the OCP and the constraints we refer the reader to our previous work in Christ et al. (2019).

## 2.11 | Local planning

The main task of the local planning module is to generate a trajectory that guides the vehicle through the local dynamic environment. Since the racing scenario requires the car to adapt quickly to new circumstances, the local planner should provide an updated trajectory every 150 ms. The trajectory should follow the global racing line from Section 2.10 whenever possible and be collision-free in multivehicle scenarios. The inputs for the trajectory generation are the map of the race track, the global racing line, the current state estimation, and the predicted behavior of the surrounding race cars. After generation, the trajectory is sent to the subsequent control module. Since the MPC-based controller reoptimizes the trajectory, the planning module also provides a collision-free corridor around the planned path serving as a constraint for the reoptimization. The state estimation is not directly used as a starting point for the planning step but is projected onto the trajectory planned in the previous step. The starting point is calculated from this projected state by interpolating along the last planned trajectory to the average calculation time. This interpolation step avoids jumping trajectories across multiple planning steps and ensures consistent motion planning.

Our trajectory planning approach is a combination of a sampling and a graph-based method. The spatiotemporal graph used is structured in layers perpendicular to a reference line using the curvilinear Frenét coordinates $s$, the longitudinal progress along the reference line, and $d$, the lateral displacement (Figure 7b). The spatial part of the graph is constructed according to Stahl et al. (2019a) and consists of spatial nodes (black points) distributed on the layers and spatial edges (gray lines) connecting the nodes. Since a search in the spatial graph alone cannot solve the combinatorial planning problem, the spatial nodes are extended by the time and velocity dimensions. Instead of using discrete values, we cover the reachable times and velocity with continuous intervals as in McNaughton et al. (2011), resulting in spatiotemporal nodes (cells of the grids). The spatio-temporal edges (red, green, and black lines) are trajectory sections connecting these spatiotemporal nodes within the layers. After generation, each spatiotemporal edge is associated with a cost.

The planning step consists of two parts depicted in Figure 7: The short-term planning step (STPS) and the long-term planning step (LTPS). While the STPS creates connecting edges from the start state (blue point) to the next graph layer, the LTPS performs a subsequent graph-search within the successive layers. Due to the update frequency of the local planning module a trajectory is not driven completely, so that the part generated by the STPS is mostly the part driven. Therefore, the main task of the STPS is to generate multiple finely planned spatiotemporal edges close to the driving limits. In contrast, the LTPS fulfills the requirement for a sufficient planning horizon by performing a coarse graph-search. This enables the planning module to react earlier to curves and other race cars. With a long planning horizon, the LTPS thus leads the STPS with a shorter horizon to ensure recursive feasibility.

Within the STPS, a set of spatiotemporal edges from the start state to the next layer is created using a polynomial approach and sampling various end conditions in Frenét coordinates. Similar to the planning concept for traffic scenarios in Werling et al. (2012), we use quartic polynomials in both lateral and longitudinal motion, as this allows for fast computation of variable and finely planned trajectories

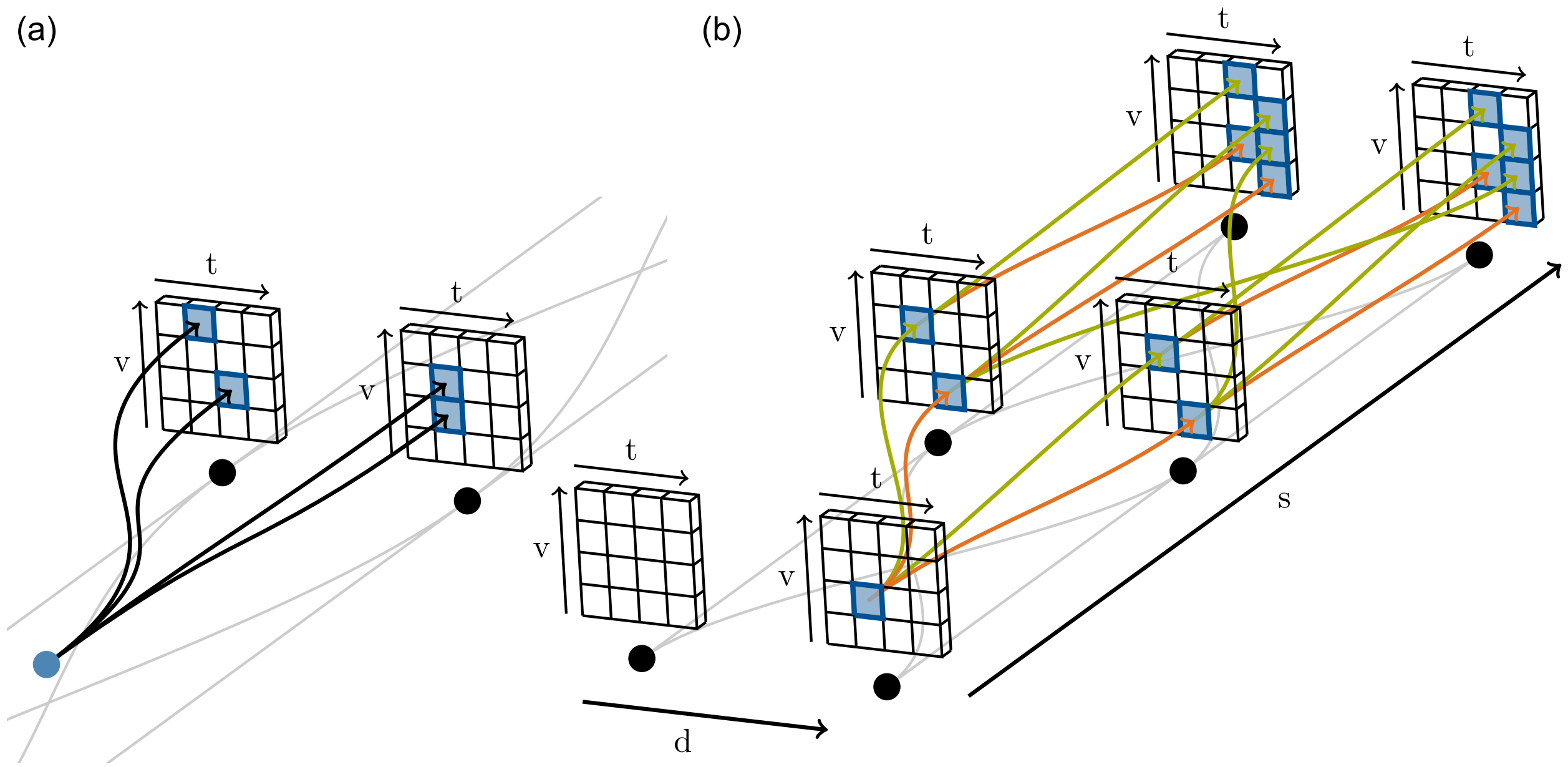

**FIGURE 7** Overview of the short-term and long-term planning steps for the local trajectory planning. (a) Short-term planning step (STPS) and (b) long-term planning step (LTPS). [Color figure can be viewed at wileyonlinelibrary.com]

for a short horizon. However, in contrast to the mentioned approach for traffic scenarios, it is necessary to embed the polynomials in the graph structure. For the edge to reach the next layer with the individual heading angle of the spatial node, the longitudinal end position and both end velocities must be specified accordingly. Besides these adaptions, the edges must also comply with the more demanding race scenario in terms of performance. Since an additional sampling of the end acceleration would increase the computation time too much, we use an iterative process to determine only one acceleration end condition for each spatial node and end velocity combination. A detailed description of the STPS and its impact on the overall race performance and safety can be found in Ögretmen et al., 2022.

Beginning at the spatiotemporal nodes connected to the start state by the edges of the STPS as exemplarily shown for one node in Figure 7b, the LTPS performs a search in the spatiotemporal graph. It generates edges connecting spatiotemporal nodes of the following layers until the desired planning horizon is reached. In general, a spatiotemporal node can be reached with multiple paths through the spatiotemporal graph, ending in the time and velocity intervals of the node (blue squares). Each path has a total cost which is the sum of the edge costs along the path beginning at the start state. Instead of expanding all reached states with edges to the next layer, only the end state of the cheapest path ending within the intervals is used as an initial state for further expansion from the considered node. This procedure follows the principle of dynamic programming and not only reduces the number of nodes by the use of intervals, but also prevents the exponential growth of the number of edges to be generated with an increasing planning horizon. As in McNaughton et al. (2011) we sample constant accelerations along the spatial edges to create the spatiotemporal edges. This is shown in Figure 7b, simplified with one acceleration (green) and one deceleration (orange) profile applied per spatial edge. Instead of an exhaustive search that proceeds from layer to layer, we perform a search based on Dijkstra's algorithm (Dijkstra, 1959) which is suited for graphs that are built during the search and often referred to as uniform-cost search. This algorithm makes it possible to go back in layers and expands only the node with the cheapest path to reach it. While the uniform-cost search retains optimality in terms of our cost function, it requires significantly fewer edges to be generated in our application, reducing the computation time by a factor of three compared with an exhaustive search. Another major advantage is that the search can be interrupted if it approaches an upper calculation time limit. In this case, we select the—at this stage—cheapest available path through the graph that satisfies the planning horizon and still obtain a suboptimal solution. A detailed description of the graph-search and the following cost function is provided in Rowold at el., 2022.

The costs of the edges determine the vehicle's behavior and must be carefully chosen to achieve safe driving on the one hand and competitive racing on the other hand. Our cost function consists of four terms, each serving a different behavior. The first term penalizes the deviation from the global racing line to reach fast laps in single-vehicle scenarios. A second term penalizes the deviation from the target speed provided by the racing line or by rules. For multivehicle racing, a prediction cost term ensures that certain lateral and longitudinal distances to other vehicles are kept and that overtaking maneuvers are initiated in time. Ellipses cover proximity regions to the predicted opponents to be avoided and provide a fast calculation for a distance measure. Finally, the fourth term penalizes the curvature for avoiding abrupt steering at high speeds. The curvature term especially comes into play in multivehicle scenarios and smoothes out overtaking maneuvers.

Besides high costs preventing spatiotemporal edges from being further considered in the search, edges can be sorted out completely. First, the edges from both the STPS and the LTPS have to be feasible in terms of maximum curvature, engine power limits, and velocity-dependent combined acceleration limits on the vehicle level so that the subsequent controller can find a reoptimized solution. Edges that exceed the limits—stored and accessed with lookup tables—are sorted out. Second, edges have to be collision-free. To determine a collision, we follow a hierarchical approach starting with oriented bounding boxes of the underlying spatial edge and ending with the exact geometry of the vehicle. Since the prediction becomes more uncertain with an increasing planning horizon, we introduce a collision-checking horizon for which the prediction is confident. Only edges that collide within this horizon are sorted out. A sufficient distance to the predictions for the rest of the trajectory, and thus recursive feasibility, is ensured by the prediction cost term. Since the edges from the STPS are mainly affected by the hard collision checks and not many behavioral options are available at this stage, there is a risk that no collision-free solution is available. In this case, we perform soft collision checks and allow colliding edges with an additional distance cost term to lead the vehicle out of the proximity region as quickly as possible.

Additionally, the local trajectory planner generates an emergency trajectory for safety reasons. Both local and emergency trajectories are sent to the control module in every planning step. The emergency trajectory decelerates on the path of the actual trajectory to eventually reach a safe state at standstill. It utilizes the full potential of the tires in terms of combined acceleration.

## 2.12 | Motion control

The motion control module is responsible for the determination of appropriate throttle, steering, and brake commands based on the planned trajectory. This includes feed-forward as well as feedback actions. The controller is structured as a three-layer concept (Figure 8a), with the highest layer utilizing a Tube-MPC with a limited friction point-mass model, an extension of a previous work (Wischnewski et al., 2021). This layer handles deviations in the position and velocity. The middle layer consists of independent proportional-integral like controllers for the lateral and longitudinal accelerations. They serve the task of matching the vehicle dynamics with the assumptions in the Tube-MPC as well as handling model inaccuracies in the utilized feed-forward control laws. The third layer adds a low-level feedback loop for the steering actuator to ensure tracking with zero steady-state error and prevent negative impacts of this subsystem on the higher-level control loops.

The second task of the motion control module is the reoptimization of the planned trajectories. Instead of applying a classical tracking control scheme, the cost function of the Tube-MPC is designed so that the lateral motion is mainly influenced by the driving tube constraints (Figure 8b) and not via a tracking target. This enables smooth driving behavior at the limits of handling, even though the graph-based local trajectory planner uses a rather coarse discretization to ensure frequent updates to the local target trajectory. However, this requires some changes to the classical MPC concept: A nominal MPC would exploit the limits aggressively, which might lead to constraint violation in the presence of disturbances or uncertainties. The proposed Tube-MPC replaces the prediction of the nominal model behavior with a set of predictions of potential uncertain outcomes (bold orange lines in Figure 8b). This leads to a closed-loop behavior that applies caution towards the end of the prediction horizon, as the optimizer requires that all constraints are fulfilled for the uncertain predictions rather than for the nominal prediction only.

The motion control software was developed using Simulink and a custom C-code integration of the numerical solver OSQP (Stellato et al., 2020). The deployment was done via code generation from Simulink and the addition of Robot Operating System 2 (ROS2) interfaces via a custom wrapper node. The software runs the main cycle of the module with a frequency of 100 Hz and handles incoming data via asynchronous callbacks. As the AV-21 does not have a dedicated real-time control ECU, we utilized real-time scheduling priorities and CPU isolation (to ensure that only specific processes or threads are scheduled on certain CPU cores) to achieve reliable execution times on the Ubuntu-based ×64 computer.

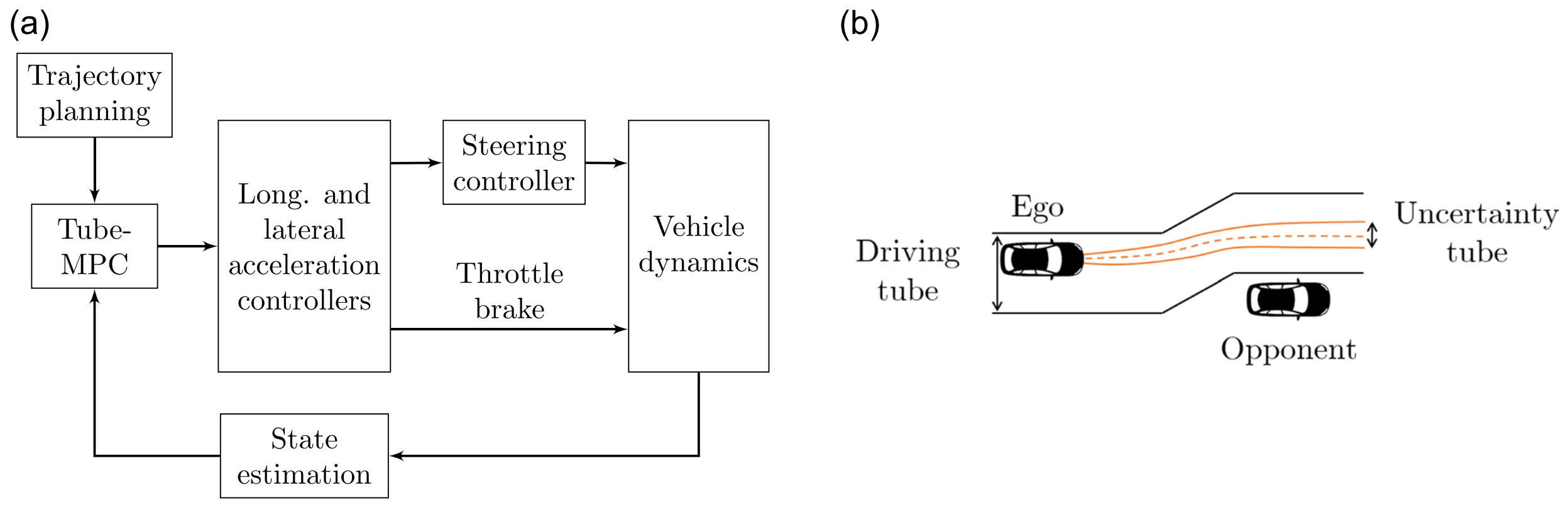

**FIGURE 8** Overview of the motion control algorithm based on Tube-MPC. (a) Internal motion control structure and (b) driving and uncertainty tubes. MPC, Model Predictive Control. [Color figure can be viewed at wileyonlinelibrary.com]

## 2.13 | Middleware and latency

Our entire software stack is based on the middleware *ROS2 Galactic*. For Data Distribution Service (DDS) implementation we rely on the open-source *Eclipse Cyclone DDS* version. For the development and deployment of the software stack, the principle of virtualization using *Docker* is applied. Here, every module corresponds to a *Docker image* that is launched via *Docker Compose* either on the vehicle or in the simulation environment. The usage of a Docker container is advantageous for deployment and versioning. The isolation especially ensures that software dependencies and requirements are not in conflict with other modules. Each container is based on an operating system (OS) base image, in our case we use Ubuntu 20.04. The running containers share a kernel with the OS of the vehicle computer. A CPU isolation was set up to ensure the computation of time-critical modules on specific cores. Using *Docker Compose* as an orchestrator each module or service can be allocated to a certain percentage of core usage.

The communication between the software stacks' modules is designed asynchronous. The default option in ROS2 for the data history is set to 10. As we work asynchronously, we do not make use of historical data and only use the most recent message. Therefore, the queue length of the message buffer is set to 1. The reliability of the quality of the service profile is set as follows: For sensor data, recent data are used at the expense of losing some, to achieve as fast as possible processing. Thus, all communication interfaces to sensors are set to best effort. Furthermore, it is necessary to ensure that the communication between modules is set to reliable, meaning that the whole data package is delivered. Due to the asynchronous character of the software, small delays in communication can occur. These delays result from communication between modules with different cycle times. Nevertheless, this feature provides a benefit. It enables flexibility within the development process of the modules, which is of importance for the overall project progress.

The software stack has no real-time behavior as the specific modules have no fixed runtime deadlines. Many modules are developed using Python, whereas time-critical algorithms are based on compiled C/C++ code. Table 2 provides an overview of the cycle time of the respective modules. The average end-to-end latency from the sensor output to the controller output results in 305.21 ms with a standard deviation of 36.40 ms for the clustering pipeline. For the RADAR pipeline an average end-to-end latency of 177.51 ms with a standard deviation of 21.33 ms occurs. Additionally, an actuator latency of 60 ms on average needs to be added to the software runtime, which can be approximated from the controller's target and actual values. All previously mentioned settings are based on the final race at the AC@CES.

# 3 | SOFTWARE DEVELOPMENT

## 3.1 | Parameter optimization and SiL testing

Important tasks in the development of autonomous race cars at the handling limits are the validation and testing of the software stack. To ensure robust vehicle behavior before testing the software on the real vehicle, two test phases are introduced: SiL and HiL testing. The SiL simulation environment is a fast way to test and validate the software stack. The test environment can be run on the developers' workstations. Some limitations arise in this environment from the fact that the perception and localization module cannot be tested. Their reliability and robustness are examined by the HiL simulator or by recordings on the real vehicle. SiL is ideally suited for software stability analysis and the investigation of predictive and planning behavior. Especially when testing vehicle behavior for rule consistency, the SiL has shown significant advantages in reducing the time investment of the developer and in increasing software reliability. Errors in the setting of race rules can be identified and corrected. The overall testing workflow concept can be seen in Figure 9.

In total, there are nine different stages in the test procedure. Usually, modified software is reviewed by the developer. As a further verification step, the collaboratively developed software is tested automatically through our own Continuous Integration/Continuous Development (CI/CD) pipeline. This pipeline checks the module for general errors in the code. In addition, a detailed test run of the entire software is performed every night. This creates a time-based performance history of the software based on defined Key Performance Indicators (KPIs). Differences, for example, in the computing time of the modules, can be detected quickly so that they can be addressed in the development process in a short time. Lap time and success rate are generally the most important objectives. Predefined scenarios are tested in the pipeline. Each scenario is evaluated in terms of successful completion. The further the vehicle gets, the better the score of the optimization run. In the event of failure, the scenario provides information about possible weaknesses in the software.

**TABLE 2** Statistical overview about the main module execution times display in ms.

| module | Mean | Std | Minimum | 25% | 50% | 75% | Maximum |
|---|---|---|---|---|---|---|---|
| Clustering | 89.05 | 22.45 | 0.33 | 85.63 | 94.09 | 100.64 | 258.76 |
| Prediction | 4.84 | 2.48 | 51.86 | 3.04 | 4.31 | 6.31 | 18.15 |
| Planning | 103.80 | 14.83 | 71.84 | 95.66 | 105.14 | 109.32 | 237.46 |
| Control | 6.43 | 2.44 | 1.43 | 4.33 | 6.80 | 8.83 | 11.28 |

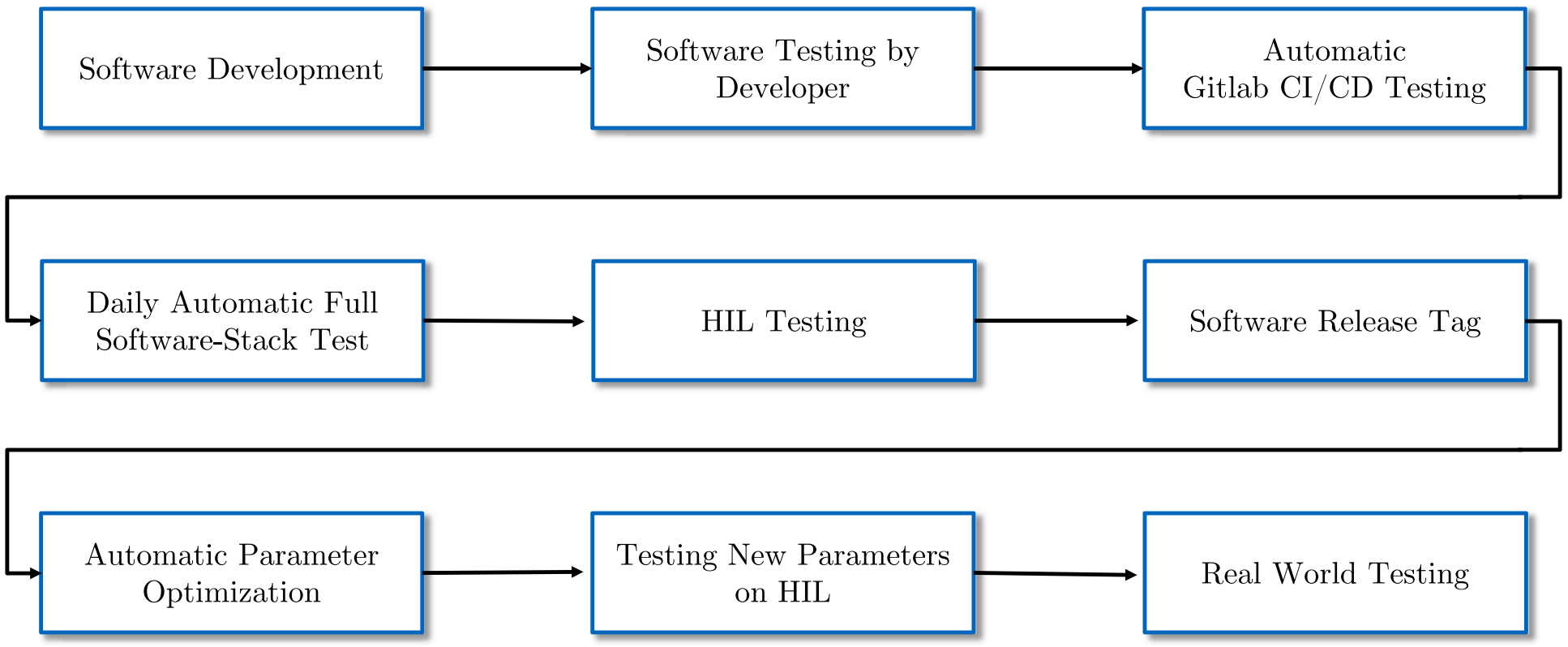

**FIGURE 9** Overall testing workflow. CD, Continuous Development; CI, Continuous Integration; HiL, Hardware-in-the-Loop. [Color figure can be viewed at wileyonlinelibrary.com]

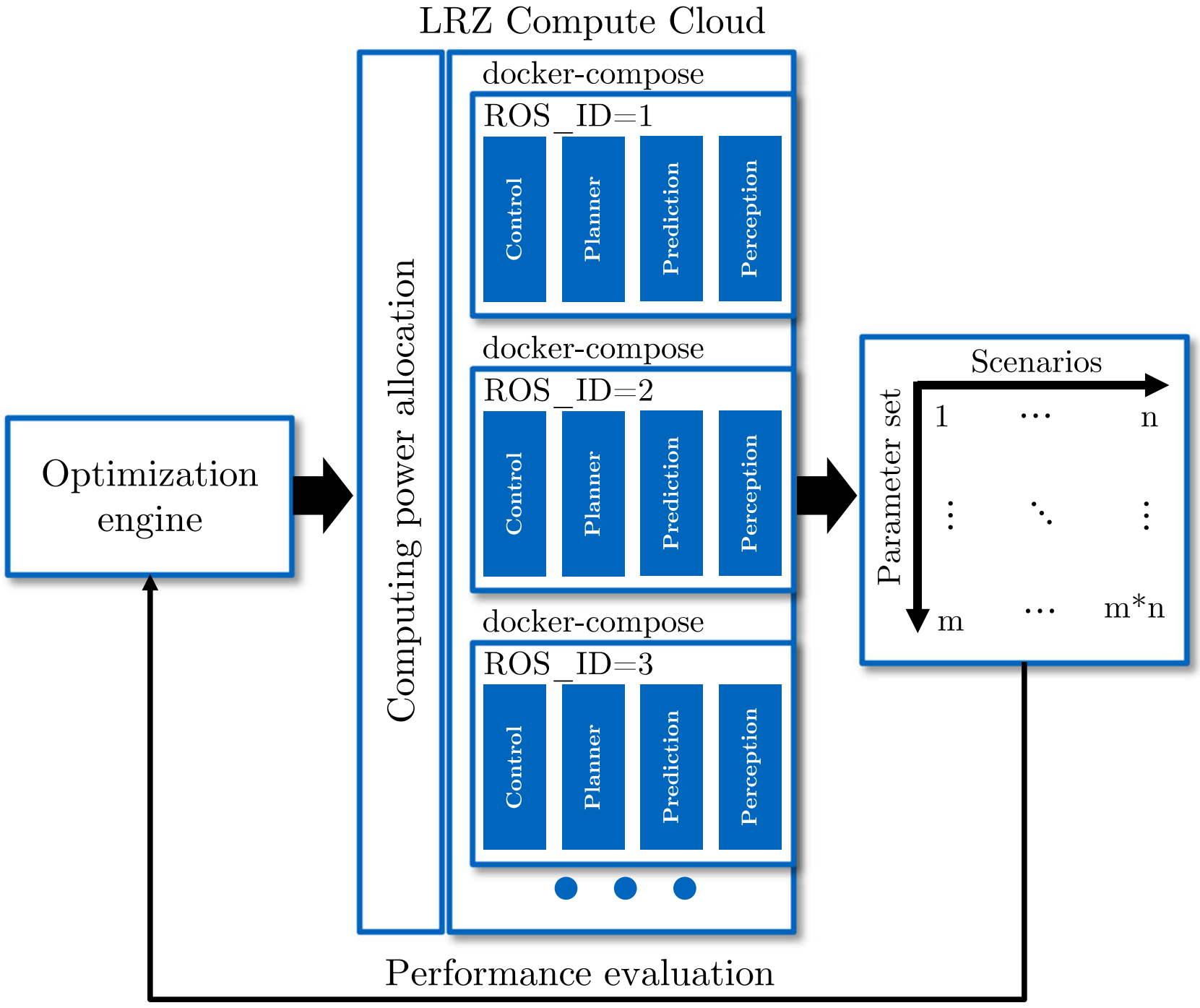

**FIGURE 10** Optimization workflow. LRZ, Leibnitz Rechenzentrum. [Color figure can be viewed at wileyonlinelibrary.com]

### 3.1.1 | Testing environment

The scenarios correspond most closely to real driving situations when opposing vehicles are on the track. Therefore, we use an additional module in the SiL to generate opponent vehicles. The generated dummy objects can follow a defined trajectory. For a more realistic simulation of the objects, it is possible to add noise to the object's perceived trajectory. A scenario catalog is defined to test the software under a variety of circumstances automatically. In addition to the performance tests, the software is also tested for emergency scenarios. For this purpose, a tool was developed that triggers an automatic error within a module in random situations. To simulate this, a single software module is disabled via Docker so that there is no further communication between that module and the rest of the software. The software must be able to bring the vehicle to a safe stop while complying with defined safety criteria.

### 3.1.2 | Automatic parameter optimization

When the software is working correctly, it needs to be optimized to meet the requirements of the racing scenarios. For this reason, we have developed an optimization tool that automatically searches for suitable module parameters. We use Nevergrad for gradient-free optimization of the parameters (Rapin & Teytaud, 2018). The tool has proven to be particularly helpful in optimizing the parameters of the planning module. Suitable cost terms of the graph planner can be determined quickly. The optimization process can be seen in Figure 10.

The optimization is executed on the LRZ Compute Cloud (Leibnitz Rechenzentrum, 2022). Different simulations can be executed simultaneously. A test-based population size adaptation method proved to be a suitable optimization algorithm since we simulate in a nondeterministic, noisy environment (Hellwig & Beyer, 2016; Liu & Teytaud, 2019). After a few hundred iterations, good results can be obtained. The gradient-free optimization leads to fast results, but due to the noisy nondeterministic environment, no global minimum can be guaranteed. The target value of the optimization is the average lap time. The misbehavior of the vehicle, such as exceeding acceleration limits, is incorporated into the lap time. If the vehicle leaves the track or causes a crash, the run is considered to have failed. In this case, the maximum distance traveled within the scenario can be used to evaluate the performance. In the next step, the parameters can be confirmed on the (HiL) simulator and the race car.

## 3.2 | HiL testing

To allow a quick and agile software development, testing, and integration workflow, a sophisticated simulation environment was developed. It enables testing of deployment-ready software independent of the vehicles and external factors. The use of such simulation environment is a crucial element for successful participation in the challenge. Beyond the SiL simulation introduced in Section 3.1, we developed a HiL simulation fulfilling the needs of autonomous racing. The setup is shown in Figure 11. This environment allows simulating one full-stack AV-21 including perception and sensors and up to nine competitors with the whole software besides perception (prediction, planning, and control). The ego-vehicle computer is a consumer desktop personal computer (PC) with specifications similar to those of the computer in the real Dallara AV-21. The other vehicles are represented by computers with a comparable CPU and no GPU. A *Speedgoat Performance* machine is responsible for calculating the vehicle dynamics of all vehicles in real-time. Therefore, a double-track model was developed and implemented in *Matlab Simulink*. The 3D scene with all vehicles and the track model is calculated and rendered on a GPU server. With the *Unity* engine, sensor models for LiDARs and cameras are realized to enable full-stack closed-loop simulation. All generated data (rosbags and internal software logs) are automatically saved on a cloud storage. A visualization and operation PC allows easy access to all components of the HiL setup and quick analysis of the runs.

To make the transition and changes from the HiL to the real car as smooth as possible, the whole ROS2 interfaces and the state machines of the AV-21 are integrated into the simulation. During switching from the real car to the HiL, the only code change that has to be made is using custom drivers for cameras and LiDARs. This also allows the integration of the race control and base station interfaces as on the actual cars.

Basic GNSS and RADAR models are integrated into the vehicle dynamics simulation and sent to the vehicle computers via user datagram protocol (UDP). The sensor drivers convert the UDP streams to ROS2 messages and publish those. For cameras and LiDARs, more detailed models were developed in the *Unity* environment. The virtual cameras are based on a pinhole camera model to render the environment and other cars from the same perspective as the real cameras. The LiDAR model is based on raycasting. Resolution and scanning patterns are adjusted to the *Luminar* LiDARs, deployed on the real vehicle, and can easily be adapted to any other LiDAR. The model incorporates noise, can handle transparent structures, and calculates intensity based on the surface color when material information is missing. The implementation of the scanning pattern also results in motion blur, which is especially important at higher speeds. Figure 12 shows a real and a synthetic point cloud from our simulator.

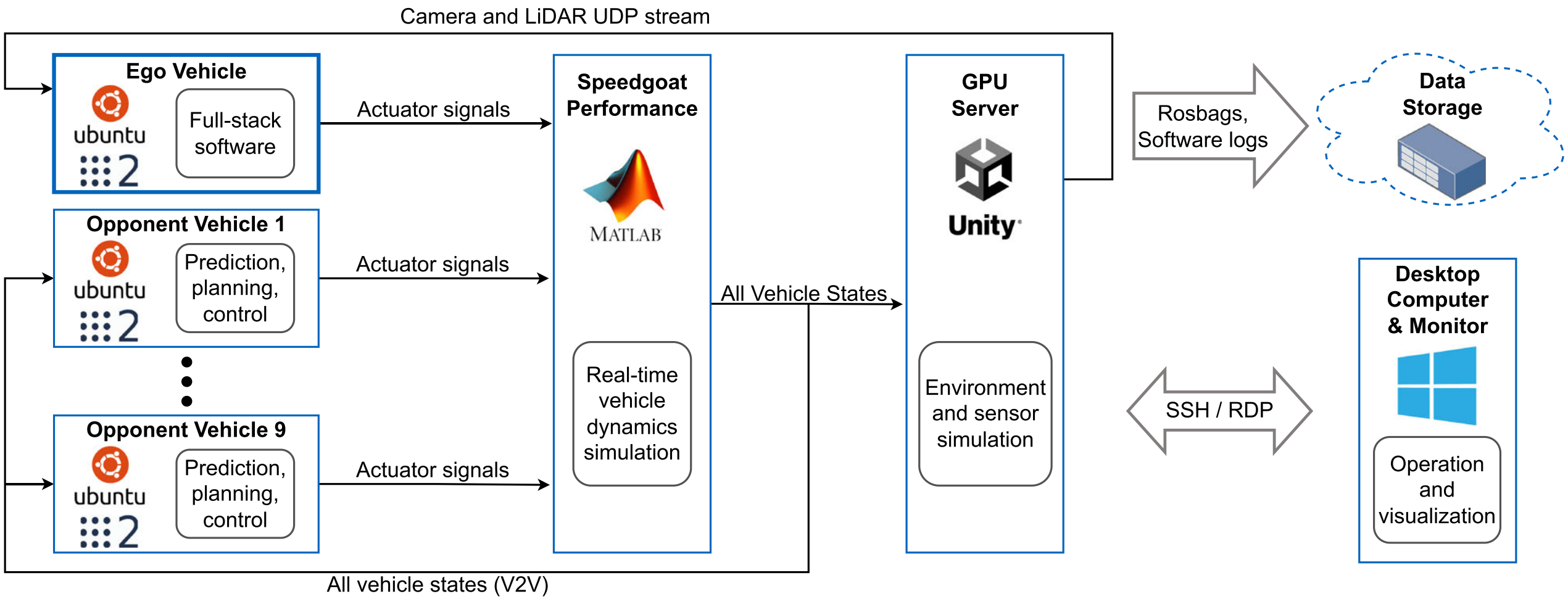

**FIGURE 11** Overview of the TUM HiL architecture. GPU, graphics processing unit; HiL, Hardware-in-the-Loop; LiDAR, Light Detection and Ranging; RDP, remote desktop protocol; SSH, secure shell; TUM, Technical University of Munich; UDP, user datagram protocol. [Color figure can be viewed at wileyonlinelibrary.com]

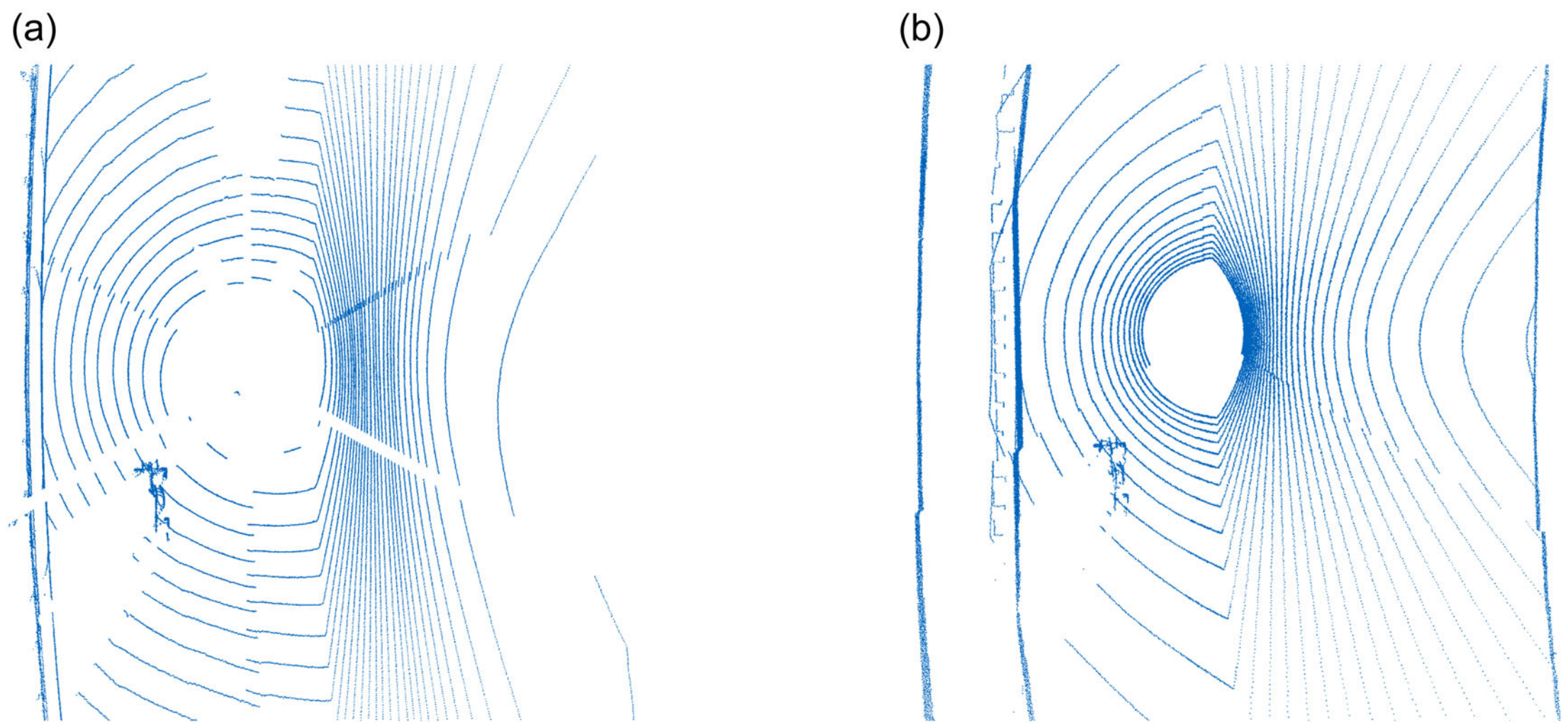

FIGURE 12 Comparison of real and synthetic point cloud. (a) Real point cloud from the AV-21 and (b) synthetic point cloud from the TUM HiL. HiL, Hardware-in-the-Loop; TUM, Technical University of Munich. [Color figure can be viewed at wileyonlinelibrary.com]

# 4 | EVENT ANALYSIS

## 4.1 | Indianapolis—Indy Autonomous Challenge

The IAC on the IMS on October 23, 2021 set out to be the first race showcasing fully autonomous race cars. In the lead-up to the race were multiple simulation challenges during the year of 2021 where multivehicle racing between the software stacks of the different teams could be shown.

The race format was as follows (Energy Systems Network, 2021): Each team can show up to two runs. Whether a second run can be performed depends on the performance in the first run. The first run is divided into a high-speed part and an obstacle avoidance part. In the high-speed part, the teams were given an out lap, two warm-up laps, and two high-speed laps. The track layout is displayed in Figure 13. Following the fast laps, two obstacles blocking opposite sides of the track must be avoided. The obstacles were placed at a random position on the start/finish straight at a longitudinal distance of 100 m. If the vehicles avoid these obstacles at a speed of 28 $ms^{-1}$, the run is considered successful. The average lap time over the two high-speed laps determined the ranking position after the first run. The three teams with the fastest averaged lap times and a successful pass of the obstacles advanced to the second and final run. In the final run, a total of four warm-up and two high-speed laps were given. The starting order was determined by the ranking of the first run. The team with the fastest averaged lap time over the two high-speed runs wins the competition.

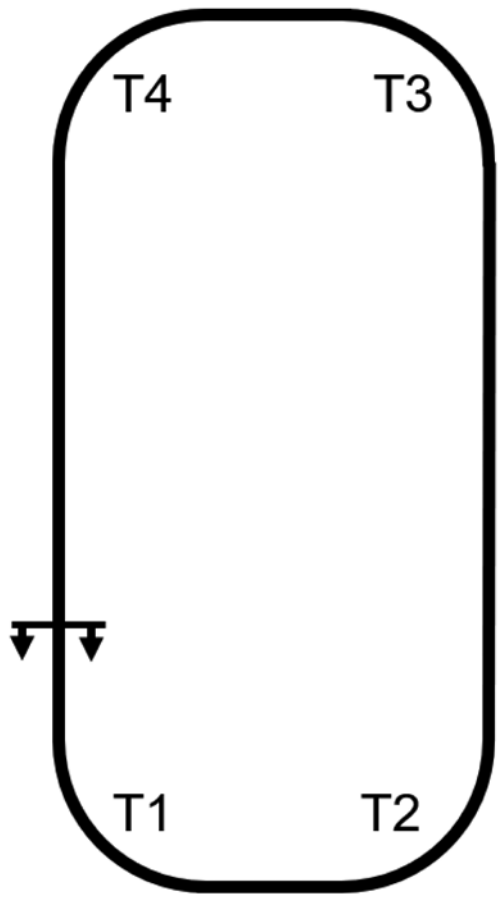

FIGURE 13 Map of the Indianapolis Motor Speedway. The track length is 4023 m with a banking of 9° throughout the turns.

Pushing into new speed ranges for autonomous vehicles brings both difficulties and learnings. When the speeds were noticeably increased during test sessions it could be observed that an important assumption of the used vehicle dynamics simulation could not be met. In the speed range up to 60 $ms^{-1}$ it proved to be difficult to get the tires up to their nominal operating temperature. Warm-up rates <8°C/lap and maximum tire core temperatures of 50°C showed that it would be difficult to reach the optimal tire temperature range of 80–100°C. This proved to be challenging since the tire data for fitting the tire model of the simulation is naturally recorded in warm conditions. The simulation as a means to estimate the vehicle performance limit is therefore subject to an unknown uncertainty in tire data. Therefore, to further increase the speed, an exploratory approach with small speed increases on the track was chosen. This example illustrates the advantages of a robust design of the algorithms to deliver a good performance even under uncertainty or disturbances. It should also be noted that despite the focus on the software side in this challenge, conventional vehicle performance aspects should not be overlooked.

With increasing speed, it could additionally be observed that at the exit and entrance of the 90° turns, there were increasing challenges in the tracking of the lateral dynamics. The lateral accelerations required by the trajectory could no longer be built up

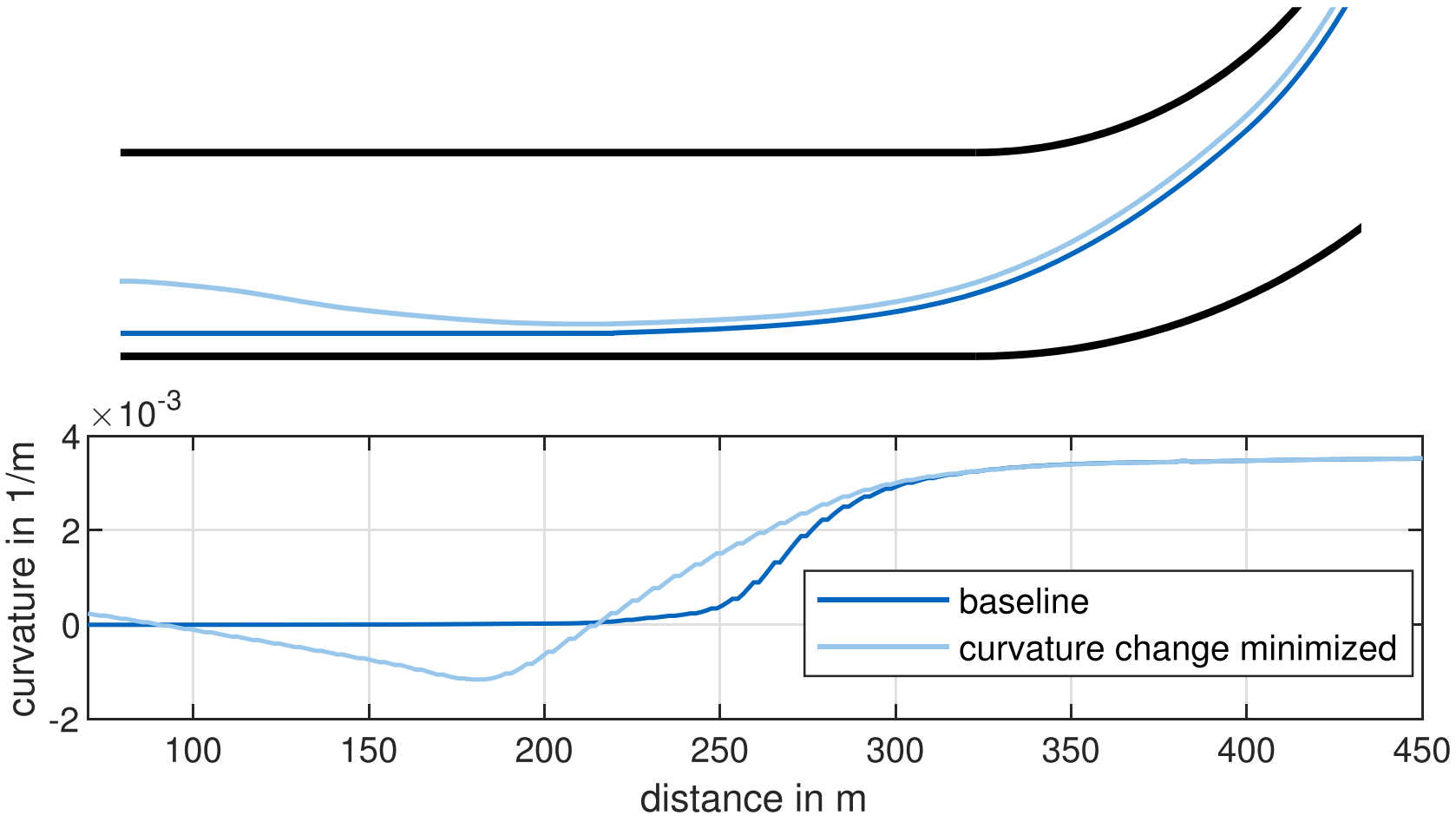

**FIGURE 14** Comparison of the racing lines. A classic racing line in blue, replaced by a curvature change minimized racing line in light blue. Due to smaller curvature gradients, the necessary rates of change for lateral acceleration and yaw rate to run the latter trajectory could thus be reduced by 38%. [Color figure can be viewed at wileyonlinelibrary.com]

and reduced at the desired rate. Analysis showed that this happened due to latencies at the steering actuator itself and on the signal path to the steering actuator, a shortcoming which was partly compensated for the Vegas event with the steering controller proposed in Section 2.12. Additionally the tire was increasingly operating in the nonlinear range, and the factors of tire run-in length and the yaw inertia of the vehicle became more relevant. To remedy this, the global trajectory of the vehicle was modified and optimized towards reduced curvature change rates. With this new global trajectory, the vehicle no longer arrives at the corner entry on the outer side of the track, but moves to the center. The vehicle then pulls outward in a swerve and finally moves towards the apex in a manner comparable to a classic racing line as shown in Figure 14. This behavior is mirrored at the exit of the turn. This is similar to a racing line often chosen by human IndyCar drivers, because it decreases the necessary yaw accelerations and therefore maximizes the combined usable tire grip for lateral acceleration.

The test period was essentially completed without the occurrence of collisions or loss of control. Internal errors could mostly be detected by the internal self-monitoring and safety stops were initiated. Once a loss of control could not be prevented by the software and an incident occurred during an attempt to increase the top speed driven up to that point. At a corner entry speed of 61 $ms^{-1}$, the car spun 360° at the exit of Turn 1 and came to a stop just before leaving the track boundary on the short chute between Turns 1 and 2. The analysis found multiple root causes:

- At the beginning of this testing day, the parameterization of the turbocharger was changed for all the cars to achieve the nominal engine performance for the event, which noticeably increased the available boost pressure. The response time of the turbocharger resulted in the throttle control starting to oscillate up against the turbocharger inertia. At low speeds, this is not noticeable because of the lower torque requests, but at speeds around 60 $ms^{-1}$, it became apparent that the throttle requests oscillated between 40% and 60%, resulting in ECU internal boost requests changes from 0% to 100%. This was strongly affecting vehicle dynamics and resulted in longitudinal acceleration oscillations of ±2 $ms^{-2}$ with a frequency of 0.5 Hz.
- Just before the spin, the input of the state estimation did not receive an update of the GPS position for several cycles. Due to cumulative integration errors of the estimation by acceleration sensor data only, a stepwise correction of the lateral error occurred when the GPS signal was received again.

The cause of the spin can be explained by the combination of the two reasons. On the one hand, the vehicle is generally closer to a vehicle-dynamically unstable state in the phases in which the drive torque is decreasing. On the other hand, the abrupt correction of the lateral error caused an additional excitation of the controller, which increased the steering angle. This led to a situation with lift-off oversteer which was favored by the increasing steering request. Since the software is conceptually not capable of counteracting oversteer but has the sole goal of fulfilling lateral deviation constraints, the described situation could no longer be solved adequately and resulted in a 360° spin. For future work, this highlighted the importance of a control module with the ability to stabilize unstable driving situations if the vehicle dynamic limit range is to be further exploited.

Another similar spin occurred on the vehicle of the contending team of *PoliMOVE*. As a consequence and in consultation with the organizer, some setup adjustments were made to all cars. The aerodynamic and mechanical balance was shifted by adjustments to wing angles and antiroll bar configuration in favor of improved grip for the rear axle to provide a greater stability reserve in comparable situations. In addition, the turbocharger parameterization was changed to ensure a linear torque delivery across the engine speed range. On the vehicle software side, the P-gain of the lower-level longitudinal acceleration controller was reduced to decrease the oscillation tendency of the accelerator pedal request.

On race day, the potential of the car could be shown. At a temperature of 12°C and cloudy conditions both runs could be finished successfully. In the first run, an average lap time of 69.7 s and an average speed of 58.4 $ms^{-1}$ could be reached. This was enough for the provisional second place and thus for a place in the final. In the second run, an average lap time of 66.2 s was achieved with an average speed of 61.5 $ms^{-1}$. The speed and lateral acceleration diagrams of the two runs can be seen in Figure 15a.

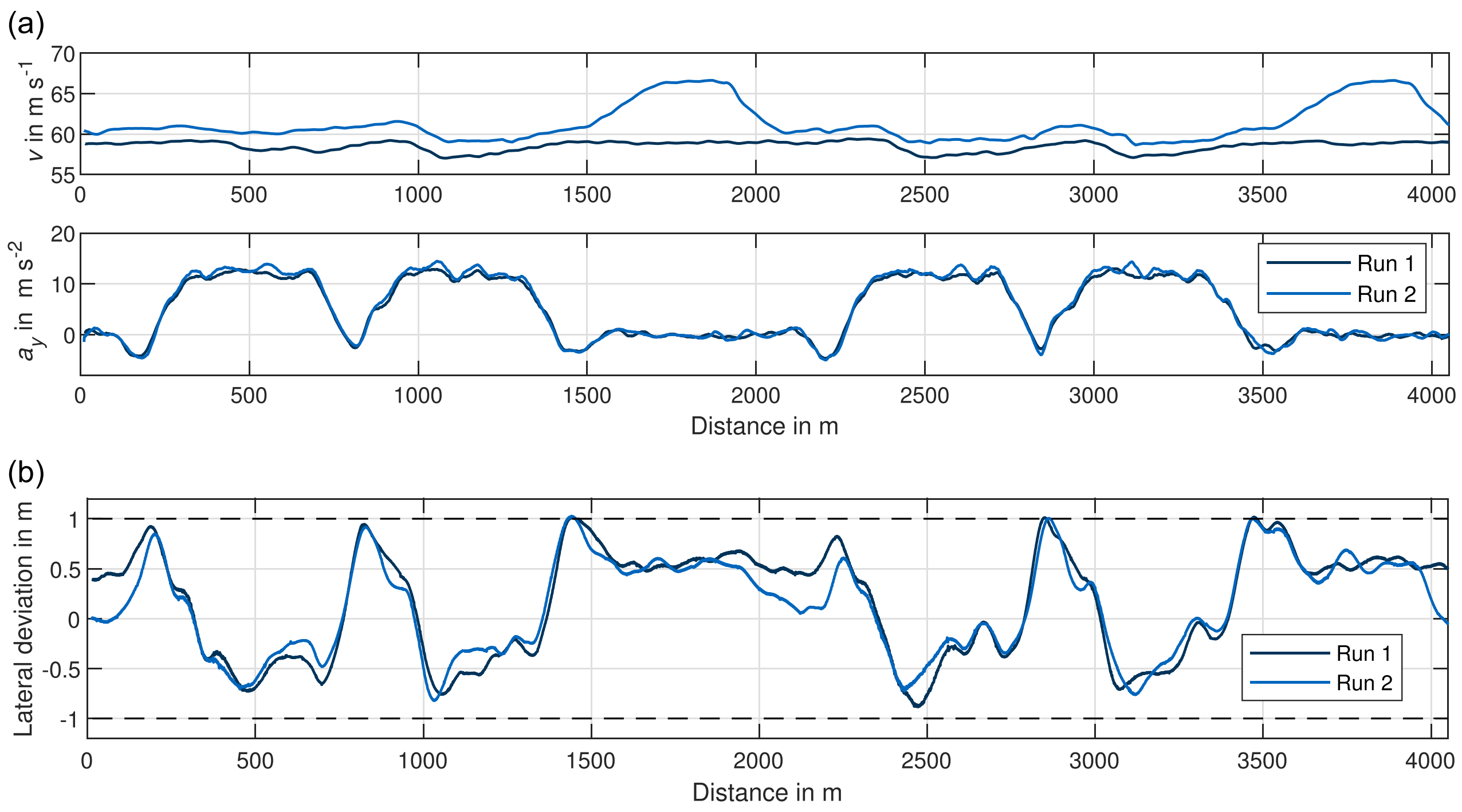

**FIGURE 15** Analysis of vehicle dynamics and controller performance at the IAC. (a) Speed and lateral acceleration: in the second run, the speed was set to 67 $m\,s^{-1}$ on the straights and 61 $m\,s^{-1}$ during the turns and (b) lateral path deviation: even though the speed in the second run was faster the lateral deviation did not increase. The reason for this was the higher gains of the low-level acceleration controller which got returned in between the runs. IAC, Indy Autonomous Challenge. [Color figure can be viewed at wileyonlinelibrary.com]

The challenge in the trade-off between speed and risk was, on the one hand, the unknown tire performance at tire temperatures below optimum. On the other hand, a limiting factor was the level of development from a hardware and software perspective achieved up to that point. This becomes particularly evident in the plot of the lateral deviation. Especially at the entry and exit of turns, lateral deviations of up to 1 m were reached. This reaches the constraints of the lateral deviation in the optimization problem of the MPC. If the lateral deviation increases to more than 1 m, the aggressiveness of the controller behavior increases significantly due to the controller design as proposed in Wischnewski et al. (2021). A lateral deviation of >1 m is still manageable, but the risk increases disproportionately above this. In Figure 15b it is shown that the maximum lateral deviation during the second run was >1.03 m.

With additional testing data, higher speeds would have been possible at the expense of higher lateral deviation. Due to the limited testing time and the unknown velocity regime, it was decided to go for a balanced compromise between velocity and risk according to the motto “To finish first, first you have to finish”. In the end this secured the first official win of the inaugural edition of the IAC which came with a Grand Prize of one Million USD.

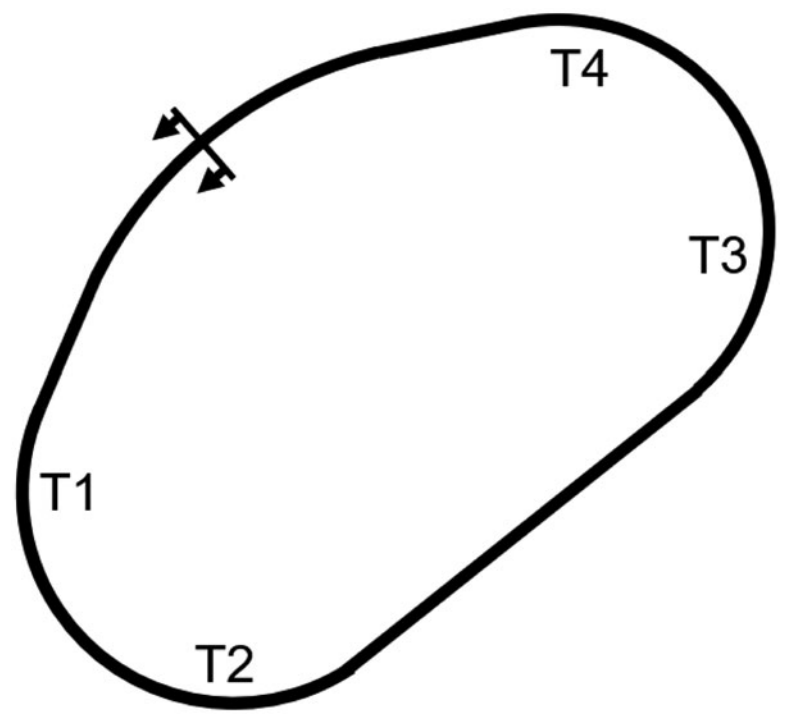

**FIGURE 16** Map of the Las Vegas Motor Speedway. The track length is 2410 m with a banking of 20° in the turns.

## 4.2 | Las Vegas—autonomous challenge at CES

The Autonomous Challenge at the CES 2022 (AC@CES), the second major event, took place on January 7, 2022 at the Las Vegas Motor Speedway (Figure 16). The event's focus was the autonomous overtaking of two competing vehicles, that is, a dual-vehicle competition. The race rules were defined as follows: The leading vehicle was the *defender* and was obligated to maintain a fixed speed with a tolerance of ±5%. Besides that, the defender had to stay on the inner side of the race track to prevent arbitrary blocking maneuvers. The trailing vehicle, the *attacker* had the task to conduct a successful overtaking maneuver against the defender within a given overtaking sector. To ensure that the attacker had a fair chance to overtake, some prerequisites must be fulfilled before the overtaking sector is entered. These were the target speed of the defender and the maximum distance threshold between attacker and defender at the start of the overtaking sector. During the overtaking the cars had to respect an exclusion zone around the vehicles, defined in a lateral and longitudinal direction. A match between two opponents followed predefined target speeds starting from 28 $ms^{-1}$ with degressive steps. The roles of attacker and defender switched after every successful overtaking maneuver. The target speed was increased to the next step as long as both vehicles were able to overtake. Hence, the team that failed first to overtake lost the match. The final event of the AC@CES was scheduled with single performance runs to evaluate the seeding based on the fastest single lap time and the main competition consisting of the described dual-vehicle challenge with two semifinals and one final run. From the technical point of view, the following aspects were our focus in preparation for the race:

- Adaption of the global map to the new race track with the challenge of a higher banking angle.
- Improved adaptation of the controller to the vehicle hardware to enable higher speeds and accelerations.
- Adjustment and fine-tuning of the perception pipeline for banking areas and fusion of multiple modalities.
- Optimization of the cost function in the local planning module for safe but aggressive overtaking behavior.

The global map and the resulting racing line had to be adjusted to the race track geometry, which provided a higher banking angle. We could benefit from the experience we made in Indianapolis and by this could release a first valid draft of the map and racing line before the first test week. However, we faced again challenges in the ego-state estimation due to our 2D representation on a 3D track. By projecting the acting forces into the plane it was possible to handle this issue. To implement this transformation, knowledge of the location-dependent road banking angle is required. Due to the significantly higher banking compared with Indianapolis, it turned out that a precise banking map is required to enable accurate localization. Since it was not possible to measure road banking directly, such a banking map could only be obtained by an iterative improvement. The residuals of the state estimation served as metrics for the evaluation. In the future, a 4+ degrees of freedom (DOF) state estimation should be used instead of a 3 DOF one to realize a more robust estimation.

To mitigate the problems described in Section 4.1 at the entry and exit of turns, the performance of the steering actuator was identified as the main cause after examining the data. In particular, a remaining control deviation in the steering angle and slow dynamics in the steering angle rates were noticed. The reasons for this were a missing steering servo which increases the load on the actuator and that the current controller of the actuator only uses a P-controller. This led to the implementation of a cascaded steering angle controller described in Section 2.12 to mitigate the mentioned problems. The performance of the steering controller is presented in Figure 17.

The tuning of the perception pipeline together with object fusion and tracking was another major task in the preparation for the dual-vehicle competition. Especially to realize a reliable detection range along the whole race track with varying banking angles was challenging as the vertical FOV of LiDAR and RADAR are quite narrow. In the case of the LiDAR we could solve this issue on the hardware side by adjusting the vertical high-density FOV along the s-coordinate on the track (Section 2.3). The setup is optimized for high sensor ranges on the straights with a tight opening angle and a comprehensive FOV with a bigger opening angle at the entrance and inside the turns to be able to detect vehicles on parallel lanes. Due to the high influence of the positioning before the overtaking maneuver, the ability to measure an object's velocity and a high sensor range of the RADAR comes into play. Due to the fact of a fixed opening angle, we optimized the RADAR perception on the software side. To cope with the high number of false positives we adjusted the filter algorithm to process the RADAR data as described in Section 2.7. Additionally, the status counter in the fusion and tracking module (Section 2.8) was adjusted to count individually per sensor depending

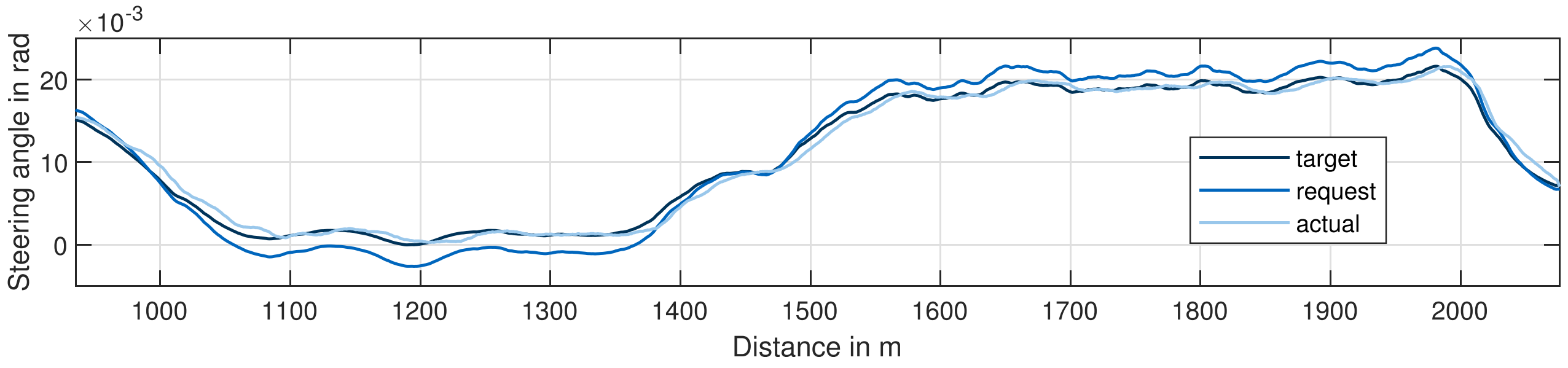

**FIGURE 17** The cascaded steering controller in real-world operation. "target" marks the steering angle that is requested by the controller. "request" denotes the signal calculated by the steering controller that is sent to the actuator. "actual" shows the sensor signal that is reported back by the actuator. It can be seen that through the implementation of the steering controller the steady-state steering angle deviations can be compensated in order that "actual" matches "target". [Color figure can be viewed at wileyonlinelibrary.com]

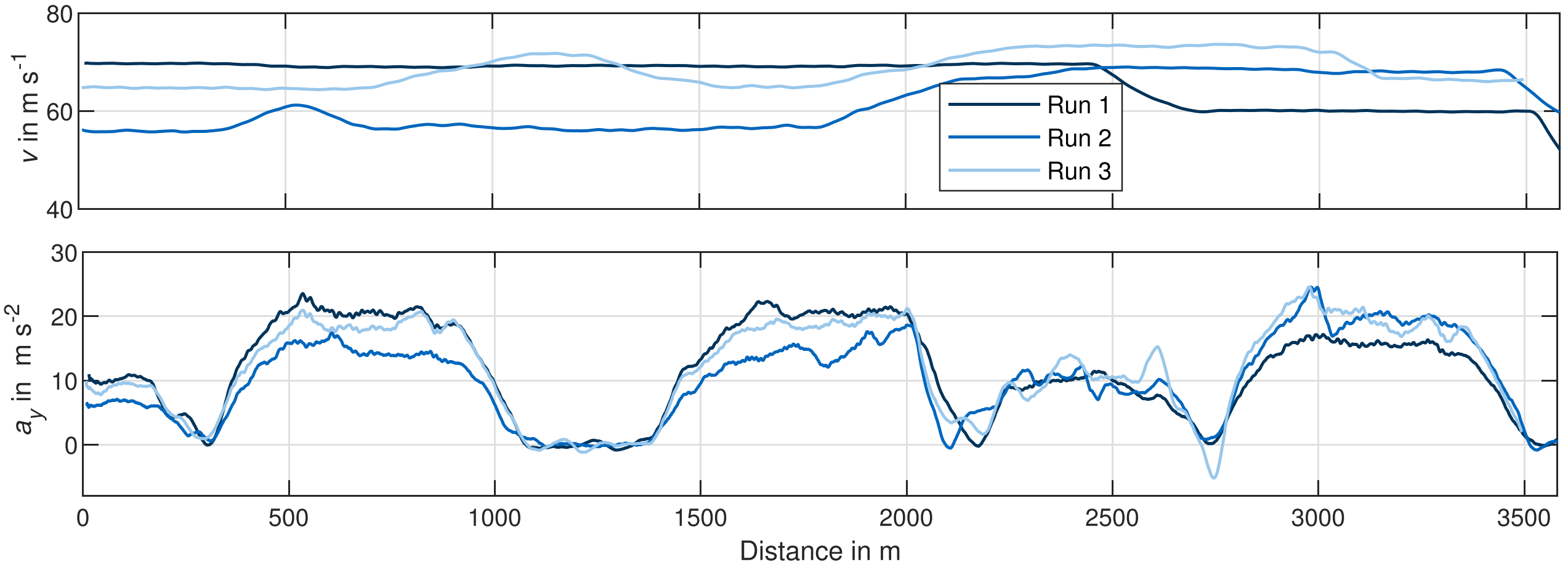

**FIGURE 18** Speed and lateral acceleration of the fastest laps and the following half lap at the AC@CES. Run 1 marks the single-vehicle qualification run with a constant speed of 70 $m\,s^{-1}$. After crossing the start/finish line at 2444 m the vehicle backs off. The second run was held with *TII EuroRacing* as a competitor. Around 500 m it can be seen that the vehicle is closing the gap to set itself up for the overtake. The overtake started in Turn 4 around 1850 m as the vehicle increased the speed. Due to a crash of the vehicle of *TII EuroRacing* at the end of the overtake, no further speed increase was achieved. The maximum lateral acceleration during the overtake is 28 $ms^{-2}$. Run 3 represents the final run against *PoliMOVE*. The vehicle closes the gap at a later stage and the process takes longer because of the higher aerodynamic resistance at higher speeds. The maximum speed achieved during this overtake is 74 $m\,s^{-1}$. [Color figure can be viewed at wileyonlinelibrary.com]

on the ego-object-distance. By this, we could track objects with higher distances more stable due to the weighted priority of the RADAR. If the object comes closer and is detectable by the LiDAR, the weighted counting is changed in the way that the effective sensitivity of the RADAR is decreased. In combination, the speed measurement and initialization of an object could be realized at high distances, but the robustness against false positives within the short range is not deteriorated.

An overview of the three event runs of the qualification, semifinal, and final is given in Figure 18. After the qualification run and a win in the semifinal due to a crash of the competing vehicle, the final event was held with the pairing of *PoliMOVE* and *TUM Autonomous Motorsport*. In the final event, we achieved a top speed of 74 $ms^{-1}$ during an overtake maneuver. However, at the next target speed step as a defender, our software triggered an emergency brake because the vehicle got unstable and oversteered on the straight at the moment when the attacker passed us. The reason for the unstable behavior was a false positive detection that led to an object predicted to cross our trajectory and a respective maneuver by the local trajectory planner. The whole combination of events will be discussed in the following.

The perception input was stable during the initialization of the overtaking maneuver, that is, when the vehicle was behind and near us, it was properly tracked over 5 s in total. Some false positives occurred, but none of them were tracked more than the single step they occurred. As the attacker advanced on the outer line at an angular position of −60° in relation to the ego-heading a high delay of 200 ms occurred in the LiDAR perception pipeline due to a high amount of reflections. Considering that the matching is distance-based it gets worse with high speed and high delay, which was the case in this step. The estimated position based on the CTRV-model had a significant yaw rate and by this, the position of the vehicle was forward integrated to the inside of the track such that the maximal matching distances did not hold anymore. As a consequence, a new object was initialized on the outer line. However, the old object was still kept in the object storage and its position was further estimated because its status counter was at the maximum value. The estimated position drifted towards the inner racing line of the ego-vehicle up to the point that the object was directly at the ego-position. The resulting prediction caused high costs due to object collision for the inner lines of the graph. Hence, an evasion maneuver was planned to the right, that is, to the outer side of the track. At this point, another factor came into play. The acceleration limits did not consider the varying banking angle along the track and especially did not consider the dependency of right and left turns. So the evasion trajectory was calculated based on the maximal positive banking angle in the turns but was indeed executed with a negative banking angle (right turn) on the straight. In combination with a slight deceleration, the tire limits at the rear axle were violated and it got unstable, which led to a spin. To sum it up, the major factors for the emergency brake were:

1. The perception delay caused by a high amount of reflection in the LiDAR pipeline.
2. The fix maximal matching distance and the fix status counter based on number of (non)detections, which did not reflect the actual uncertainty depending on the object's speed and perception delay.
3. The missing spatial dependency of the acceleration limits in the trajectory planning to evaluate the driveability of trajectories more precisely.

4. The missing consideration of the curvature rate (correlation with lateral jerk) in the cost function of the trajectory planning to prevent high curvature changes.

All the factors were known before but were approved to resolve trade-offs with other performance indicators. It becomes obvious that the sum of small weak spots can result in such a situation if the car is close to its handling limits. However, the collected data of sensors and ego-vehicle's interactive behavior are of high value to further improve the software as the insights serve as specific starting points for future development steps.

## 5 | DISCUSSION

### 5.1 | Evolutionary software stack development

The interdisciplinary research group *TUM Autonomous Motorsport* started its participation in autonomous racing events in early 2018 with a demonstration of high-speed single-vehicle behavior on the Berlin Formula-E circuit in conjunction with Roborace. Afterward, the software stack and simulation capabilities were extended to multi-vehicle scenarios and close-to-human lap-time performance on the same vehicle platform in 2019. On the basis of these developments and achievements, the software stack displayed in this paper for participation in the IAC competitions was created. The evolutionary development of the software stack provided the chance to reuse modules and rethink some software components that needed to be developed from scratch again. Furthermore, the evolutionary development provided the change to replace old software modules with more powerful ones. For example, while in the Roborace competition, classical control approaches for path and velocity tracking were chosen, in the IAC, the more advanced technology of Tube-MPC was used. Building upon the previous knowledge allows the comparison between methods, evaluation of their performance, the integration of more aggressive algorithms that can handle the car at the dynamical limits and ultimately leading to the software stack displayed in this paper.

### 5.2 | Lessons learned

#### 5.2.1 | Autonomous software design guidelines

This software stack was designed to handle multivehicle racing scenarios with various opponent vehicles and is scalable depending on the available computational resources. The primary design guidelines were: First, a modular and comprehensive software architecture that can handle racing and other autonomous driving challenges. Second, early and extensive full-stack testing in simulation to determine the influence and sensitivity of particular algorithms on the overall software level and fast iteration through a solid CI and testing framework. Third, robust real-world performance via the proactive consideration of uncertainties and failures in each algorithm. In addition to achieving those goals, we benchmarked the proposed architecture and the developed algorithms under realistic conditions, which led to several important insights shaping our current and future research strategy. The holistic approach to these research challenges allowed us to generate further insights and learnings.

#### 5.2.2 | Autonomous racing as an operational design domain (ODD)

An important design decision for many developers is the specification of an ODD. While it allows focusing on specific aspects of the problem, this strategy often leads to a crucial pitfall: Many algorithms are prone to complete failure if the assumptions made within the ODD are slightly violated. Examples of this are hard constraints in motion planning or model-predictive-control algorithms. While this does not lead to severe issues in isolated applications and benchmarks, the inherent uncertainties (either caused by sensor input noise, inaccurate model assumptions, or numerical issues) propagating through the software stack will almost certainly lead to frequent issues with algorithms crashing or becoming infeasible. Therefore, it is of paramount importance to understand the behavior of the algorithms when the ODD is violated to a certain extent and ensure that the response remains reasonable, for example, via the introduction of soft constraints. The violation of predefined domain assumptions must also be considered during the concept phase by choosing generic and robust algorithms. Even though a lower module performance in contrast to more specific and overfitted algorithms might occur, at first sight, the robustness pays off in the long run when it comes to real-world applications with the mentioned uncertainties and overall software integration. Additionally, a valid safety concept to handle module failures is essential to ensure safety on the one hand and on the other hand to enable the integration of new features while still having a fallback option in case of ODD violations.

#### 5.2.3 | Model fidelity versus software performance

There exists a counterintuitive relation between increasing model fidelity and the increase in overall software performance. The low complexity of the algorithms leads to a software stack that has difficulties adjusting to the behavior of other vehicles on track or other deviations from the internal assumptions. There are two ways to counteract this issue: The first, probably the more common, is the introduction of more complex models of reality. However, this almost certainly leads to an increase in computational costs and, therefore, a decrease in update rate. This might lead to worse overall performance, even though the utilized model improves accuracy as the higher latency strongly influences the opportunity to react adequately in dynamic situations. The second strategy is to keep the complexity and related computational costs of particular models low and optimize the overall software latency. Upgrades to more complex models are strictly prioritized by the model's influence on the overall software performance, that is, bottlenecks must be identified a priori. This strategy has proven promising during our development; however, it is much harder to measure or evaluate as it strongly depends on the test cases and performance indicators. This finding emphasizes the importance of overall software stack performance rather than measuring KPIs of individual algorithms. Consequently, early integration and standardized testing are of high relevance to ensure the compatibility of new features and to track the progress of the overall software performance.

### 5.2.4 | Data-driven algorithms

Data-driven algorithms are prone to a *chicken-and-egg problem*. Their use relies on the availability of data, which is hard to acquire in an autonomous vehicle when the acquisition requires the desired capability to be available. While this issue is often circumvented via human test drivers, drones, or other data collection equipment, this will challenge especially research groups, and smaller companies as their access to realistic data is limited. A potential way of tackling this issue is the gradual introduction of data-driven strategies with increasing capabilities of the software stack. When the software stack initially uses classical algorithms, the utilization of data-driven algorithms can increase and improve the software stack with increasing maturity and data availability. For this, the design of a modular architecture is required to continuously integrate new features but still be able to run the complete software for testing and data collecting purposes. In addition, synthetic data generation in a versatile simulation framework of sensor and vehicle dynamics simulation is crucial. If these techniques are used, training of deep-learning models and parameter tuning of complex algorithms can take place a priori to real-world operation and can continue to be conducted during the development phase.

## 6 | CONCLUSION AND FUTURE WORK

This paper presented the autonomous racing software stack developed by TUM Autonomous Motorsport. We displayed the content of the individual software modules capable of multivehicle racing at high speeds and high accelerations. It was demonstrated that the software drives close to the Dallara AV-21s limit by peaking at around 270 $\mathrm{km}h^{-1}$ and 28 $ms^{-2}$. Furthermore, by developing a dedicated testing and development pipeline, we created a robust and advantageous software that is tested in various simulations and real-world racing competitions. The experiences and learnings during the application of this software stack at the IAC allowed us to identify crucial further research directions to enable safe autonomy in the future:

First and foremost, the transfer of algorithms and knowledge among different domains of autonomy has to be improved. While autonomous racing with one or two vehicles is a reasonable proving ground, we see a strong need to increase the complexity of these challenges to align with the problems faced in urban and highway scenarios. An essential part of this is racing more than two vehicles simultaneously to prove the algorithms' interaction awareness and scalability. As this strategy increases the risk of vehicle damage, we identify a strong need for improved open and freely accessible resources for virtual development. Open-source projects like CARLA are promising but have not been adopted on a wide scale in these large competitive projects.

Second, the thorough handling of uncertainties (and their multimodal nature) through the whole software stack will be an essential part of increasing the safety of autonomous vehicles. While promising approaches for independent algorithms such as object detection, prediction, and planning are available, these parts must be combined and evaluated as a full software stack. This will be especially challenging from a computational complexity point of view. It seems necessary that these holistic approaches employ significant parallelization of their workload, either via an increased number of CPU cores or the employment of GPU-based calculations.

Third, the development workflow has to be considered as an active research direction rather than an industrialization challenge. The safe and efficient deployment of autonomous vehicles in various applications will depend heavily on the ability of companies to iterate quickly to generate learnings on their approach while having safety requirements complying with all guidelines and highest standards. This especially includes the software development workflow, requirements specification, testing scenario design, and holistic tracking of algorithm performance from a virtual, single algorithm level up to the full software stack.

Lastly, it remains to keep working towards a free racing format and increase the complexity of the race situations. The era of autonomous racing is relatively new, but we already see the benefit of gaining new insights for research and development, which are transferable to further autonomous applications. The goal is to enable multivehicle races on road courses and oval tracks. The required rules should be minimized such that they ensure the basics of safety and fairness but should support a dynamic, interactive, and free racing style. Both the rule format and the track selection led to the development of generic and robust software stacks highly correlated to software for autonomous driving on public roads.

**ACKNOWLEDGMENTS**

We want to thank the Indy Autonomous Challenge organizers, Juncos Hollinger Racing, and all other participating teams for the countless efforts to make the Indy Autonomous Challenge and all of those experiments with multiple full-scale autonomous racing vehicles possible. Furthermore, this project was made possible with the generous support and contributions of the basic research funds of the Technical University of Munich and several private donors and sponsors. Open Access funding enabled and organized by Projekt DEAL.

## DATA AVAILABILITY STATEMENT

The data that support the findings of this study are openly available in TUM—Institute of Automotive Technology at https://github.com/TUMFTM.

## ORCID

*Phillip Karle* http://orcid.org/0000-0003-3223-6969